\documentclass[11pt, letterpaper]{article} 
\usepackage{hyperref}
\usepackage{url}
\usepackage{amsmath,amssymb,amsthm, amsfonts}
\usepackage[T1]{fontenc}
\usepackage{mathtools}
\usepackage{xcolor}
\usepackage{enumitem, comment, xifthen}
\usepackage{graphicx}
\usepackage{mathabx} 
\graphicspath{{figs/}}
\DeclarePairedDelimiter{\vecnorm}{\lVert}{\rVert}

\newcommand{\twonorm}[2][]{\vecnorm[#1]{#2}_{2}}

\makeatletter
\def\letterdef#1#2#3{\def\letterdef@##1{\expandafter\def\csname #1\endcsname{#2}}%
  \letterdef@@#3{?\@car{}}\@nil}
\def\letterdef@@#1{\@gobble#1\letterdef@{#1}\letterdef@@}
\makeatother
\newcommand{\R}{\mathbf R} 
\letterdef{c#1}{\mathcal{#1}}{ABCDEFGHIJKLMNOPQRSTUVWXYZ}
\letterdef{b#1}{\mathbb{#1}}{ABCDEFGHIJKLMNOPQRSTUVWXYZ}
\let\defn\coloneq

\newcommand{\e}{\mathrm{e}}

\newcommand{\1}{\mathbf 1} 
\let\ones\1
 
\let\epsilon\varepsilon
\newcommand{\eps}{\varepsilon}
\newcommand{\iid}{\mathrm{i.i.d.}}
\newcommand{\simiid}{\stackrel{\iid}{\sim}} 

\makeatletter
\newcommand{\E}{\operatorname*{\mathbf{E}}\ilimits@}
\makeatother
\makeatletter
\renewcommand{\P}{\operatorname*{\mathbf{P}}\ilimits@}
\makeatother
\makeatletter
\long\def\@makecaption#1#2{
        \vskip 0.8ex
        \setbox\@tempboxa\hbox{\small {\bf #1.} #2}
        \parindent 1.5em 
        \dimen0=\hsize
        \advance\dimen0 by -3em
        \ifdim \wd\@tempboxa >\dimen0
                \hbox to \hsize{
                        \parindent 0em
                        \hfil 
                        \parbox{\dimen0}{\def\baselinestretch{0.96}\small
                                {\bf #1.} #2
                                } 
                        \hfil}
        \else \hbox to \hsize{\hfil \box\@tempboxa \hfil}
        \fi
        }
\makeatother

\newcommand{\dimension}{d}
\newcommand{\numcomponents}{m}
\newcommand{\promptlen}{k}
\newcommand{\Normal}[2]{\mathsf{N}(#1, #2)}
\newcommand{\MSE}{\mathrm{MSE}}
\newcommand{\mixturedist}{\pi}
\DeclareMathOperator{\RAW}{\mathsf{RAW}}
\DeclareMathOperator{\copyover}{\mathsf{copy\_over}}
\DeclareMathOperator{\copydown}{\mathsf{copy\_down}}
\DeclareMathOperator{\mulop}{\mathsf{mul}}
\DeclareMathOperator{\affineop}{\mathsf{aff}}
\DeclareMathOperator{\softmaxop}{\mathsf{soft}}
\DeclareMathOperator{\scaledagg}{\mathsf{scaled\_agg}}
\DeclareMathOperator{\sigmoid}{\mathsf{sigmoid}}

\DeclareMathOperator{\divop}{\mathsf{div}}
\DeclareMathOperator{\movop}{\mathsf{mov}}

\newcommand{\evenindices}{\cI_{\rm even}(\promptlen)}
\newcommand{\oddindices}{\cI_{\rm odd}(\promptlen)}
\newcommand{\clustermatrix}{\cW^\star}
\newcommand{\dimattention}{d_{\rm att}}
\newcommand{\numheads}{{n_{\rm heads}}}
\newcommand{\dimff}{d_{\rm ff}}
\newcommand{\QueryMat}{W^{\rm Q}}
\newcommand{\ValueMat}{W^{\rm V}}
\newcommand{\KeyMat}{W^{\rm K}}
\newcommand{\CombineMat}{W^{\rm C}}
\newcommand{\FFInMat}{W^{\rm in}}
\newcommand{\FFOutMat}{W^{\rm out}}
\DeclareMathOperator{\softmax}{softmax}
\DeclareMathOperator{\layernorm}{\lambda}
\DeclareMathOperator{\nonlin}{\sigma}

\newcommand{\ie}{\textit{i}.\textit{e}., }

\DeclareMathOperator{\argmin}{\arg\min}
\makeatletter
\newcommand*{\transpose}{%
  {\mathpalette\@transpose{}}%
}
\newcommand*{\@transpose}[2]{%
  \raisebox{\depth}{$\m@th#1\intercal$}%
}
\makeatother
\newcommand{\T}{\transpose}

\theoremstyle{plain}

\newtheorem{theo}{Theorem}

\newtheorem{lem}{Lemma}
\newtheorem{prop}{Proposition}
\newtheorem{cor}{Corollary}

\theoremstyle{definition} 

\newtheorem{nota}{Notation}
\newtheorem{de}{Definition}
\newtheorem{exa}{Example}
\newtheorem{as}{Assumption}
\newtheorem{alg}{Algorithm}

\newcommand{\btheo}{\begin{theo}}
\newcommand{\bde}{\begin{de}}
\newcommand{\ble}{\begin{lem}}
\newcommand{\bpr}{\begin{prop}}
\newcommand{\bno}{\begin{nota}}
\newcommand{\bex}{\begin{exa}}
\newcommand{\bcor}{\begin{cor}}
\newcommand{\spro}{\begin{proof}}
\newcommand{\bas}{\begin{as}}
\newcommand{\balg}{\begin{alg}}

\newcommand{\etheo}{\end{theo}}
\newcommand{\ede}{\end{de}}
\newcommand{\ele}{\end{lem}}
\newcommand{\epr}{\end{prop}}
\newcommand{\eno}{\end{nota}}
\newcommand{\eex}{\end{exa}}
\newcommand{\ecor}{\end{cor}}
\newcommand{\fpro}{\end{proof}}
\newcommand{\eas}{\end{as}}
\newcommand{\ealg}{\end{alg}}

\theoremstyle{plain}

\newtheorem{theos}{Theorem}
\newtheorem{props}{Proposition}
\newtheorem{lems}{Lemma}
\newtheorem{cors}{Corollary}

\theoremstyle{definition}
\newtheorem{exas}{Example}
\newtheorem{algs}{Algorithm}
\newtheorem{asss}{Assumption}
\newtheorem{defns}{Definition}

\newcommand{\btheos}{\begin{theos}}
\newcommand{\etheos}{\end{theos}}
\newcommand{\bprops}{\begin{props}}
\newcommand{\eprops}{\end{props}}
\newcommand{\bdes}{\begin{defns}}
\newcommand{\edes}{\end{defns}}
\newcommand{\blems}{\begin{lems}}
\newcommand{\elems}{\end{lems}}
\newcommand{\bcors}{\begin{cors}}
\newcommand{\ecors}{\end{cors}}
\newcommand{\bexs}{\begin{exas}}
\newcommand{\eexs}{\end{exas}}
\newcommand{\balgs}{\begin{algs}}
\newcommand{\ealgs}{\end{algs}}
\newcommand{\bass}{\begin{asss}}
\newcommand{\eass}{\end{asss}}

\usepackage[margin=1.15in]{geometry}
\usepackage{hyperref}
\usepackage{url}
\usepackage{float}
\usepackage{forest}
\usepackage{algorithm}
\usepackage{algpseudocode}
\usepackage{natbib}

\begin{document}
\begin{center}
{\bf{\Large 
  Transformers can optimally learn regression mixture models}} \\
  \vspace*{.2in}
{\large{
 \begin{tabular}{cccc}
  Reese Pathak$^{\diamond}$ &
  Rajat Sen$^{\dagger}$ &
  Weihao Kong$^\dagger$ & 
  Abhimanyu Das$^\dagger$ 
 \end{tabular}\\[2ex]
\begin{tabular}{c}
  UC Berkeley, EECS$^{\diamond}$ \\
  Google Research$^{\dagger}$
 \end{tabular}
}}
\end{center} 
\medskip
\begin{abstract}
Mixture models arise in many regression problems, but most methods have 
seen limited adoption partly due to these algorithms' highly-tailored and model-specific nature. 
On the other hand, transformers are flexible, neural sequence models 
that present the intriguing possibility of providing general-purpose 
prediction methods, even in this mixture setting. 
In this work, we investigate the hypothesis that transformers can learn an optimal predictor for mixtures of regressions. 
We construct a generative process for a mixture of linear regressions 
for which the decision-theoretic optimal procedure is given by data-driven exponential 
weights on a finite set of parameters. 
We observe that transformers achieve low 
mean-squared error on data generated via this process.
By probing the transformer's output at inference time, we also show that transformers typically make predictions that are close to the optimal predictor. 
Our experiments also demonstrate that transformers can learn mixtures of regressions in a sample-efficient fashion and are somewhat robust to distribution shifts. 
We complement our experimental observations by proving constructively that the decision-theoretic optimal procedure is indeed implementable by a transformer.
\end{abstract}

\section{Introduction}
\label{sec:intro}
In several machine learning applications---federated learning~\citep{wang2021field}, crowd-sourcing \citep{steinhardt2016avoiding} and recommendations systems~\citep{wang2006unifying}---data is collected from multiple sources. 
Each source generally provides a small batch of data: for instance in recommendation systems, a user can provide a source of rating data on a subset of items 
that she has encountered. Such batches, on their own, are often too small to learn an effective model for the desired application. On the other hand, by pooling many batches together, improvements can typically be made in the quality of the predictors that 
can be learned. 

An issue with this ``pooling'' approach is that if it is done carelessly, 
then the models which are learned may lack personalization~\citep{ting1999learning}. 
For instance, in a recommendation system, such an approach could yield a model that 
selects similar actions for dissimilar users. A better approach, however, is to model 
the problem as a \emph{mixture} of distributions: for instance, we can 
model the sources as arising from $m$ subpopulations, assuming that
sources arising from the subpopulation have similar underlying distributions~\citep{kleinberg2004using}. The sources from a single subpopulation can then be aggregated for the purposes of learning. For instance, in the recommendation systems example, users in the same subpopulation might be identified as having 
similar preferences and tastes for item genres. 

A supervised learning formulation of the above setup is that the sources arise from a subpopulation indexed by an integer $i \in [\numcomponents] \defn \{1,2,\cdots,\numcomponents\}$. Additionally, assume that within each subpopulation the input-output pair $(x, y)$ follows 
a relation of the form $y = f^{\star}_i(x) + \eta$, where $\eta$ is a zero-mean noise, 
and $x \in \R^\dimension$. 
A batch of i.i.d data from such a source can be represented as 
$\{(x_j, y_j)\}_{j=1}^{\promptlen}$ where $\promptlen$ is the batch size. 
Given many such batches, each having examples only from one source, the task is to learn the functions $\{f^{\star}_{i}\}_{i=1}^\numcomponents$ well enough to make 
good predictions on another input, sometimes also referred to as a \emph{query}, $x_{\promptlen+1}$. For instance, given the past ratings of an user, we should be able to determine their subpopulation well enough to infer their 
preferences on an unseen item.

The simplest version of the formulation above additionally imposes the 
assumption that the functions $f^\star_i$ are linear: $f^{\star}_{i}(x) = {\langle w^{\star}_i, x_i \rangle}$. This setting has been studied theoretically in~\citep{kong2020meta, jain2023linear}. \citet{kong2020meta} introduced the problem and designed an algorithm in the setting where there are as many as $O(d)$ batches 
with size $\promptlen = O(1)$, and fewer medium-sized batched of size $\promptlen = O(\sqrt{m})$. However, that work imposed strong assumptions on the covariate distribution, which lead to the paper \citet{jain2023linear}, where these 
assumptions were relaxed. This latter work proposes a different algorithm that even allows covariate distributions to vary among subpopulations. Nonetheless, their algorithm needs to know problem parameters, such as a  $L^2-L^4$ hypercontractivity, a spectral norm bound on the covariance matrix, the noise level, and the number of subpopulations $m$. It is not clear how these algorithms will fare with model misspecification, or if they could be extended to applications like federated learning where it is unlikely that the correct model is linear, and distributed algorithms are required.

In this work, we ask the question: \textit{Is there a deep learning architecture that can be trained using standard gradient decent, yet learns mixture models from batched data and can leverage small batches from a source to make predictions for its appropriate subpopulation?} If so, this would allow us to solve these type of mixture problem without needing highly specialized algorithms that could potentially be brittle with respect to knowing the correct form of the mixture model. Moreover, standard (stochastic) gradient descent would naturally extends to distributed training, using standard techniques from federated learning~\citep{wang2006unifying}. 

A natural candidate to address this question is the widely used transformer architecture~\citep{VasEtAl2017}. Motivated by their immense success in NLP~\citep{radford2019language}, computer vision~\citep{han2022survey} and in context learning abilities demonstrated by large models~\citep{chowdhery2022palm}, several recent works have been aimed to analyze whether transformers can learn algorithms~\citep{MaEtAl2023, garg2022can, von2023transformers}. These papers train decoder only transformers using prompts of the form $(x_1, f(x_1),\cdots,x_j, f(x_j), \cdots x_k, f(x_k), x_{k+1})$ where the task is to predict $f(x_{k+1})$ after seeing the portion of the prompt preceding it. These papers show empirically that when $f$ is sampled from a class of linear functions, then transformers learn to perform linear regression in-context. \citet{MaEtAl2023} also show that transformers can represent gradient descent in the case of linear regression provided through a construction. 

\paragraph{Our contributions:} 
Motivated by the above papers, we investigate whether transformers can solve the problem of learning mixture models from batched data. Our contributions are as follows:
\begin{itemize}
    \item We demonstrate that transformers can learn mixtures of linear regressions by training on such mixture data and exhibiting near-Bayes-optimal error at inference time.
    \item We strengthen this observation by proving constructively that transformers can implement the optimal method for the mixture distribution on which the transformer was trained. 
    \item Our experiments show that transformers are sample-efficient: the transformers' performance is similar (or better) than model-specific methods, when fixing the training set size. 
    \item We evaluate certain inference-time metrics that capture the nearness of predictions made by the transformer versus another predictor. We show these metrics are smallest when taking the comparator to be the decision-theoretic optimal method, thereby further corroborating the hypothesis that transformers optimally learn mixtures of regressions. 
    \item We suggest that transformers tolerate ``small'' distribution shifts by investigating transformers' performance on both covariate and label shifts 
    to the mixture model.
\end{itemize}

These contributions, taken together, are evidence that transformers can optimally, efficiently, and robustly learn mixtures of linear regressions. 

\paragraph{Related work:} 
The related work can be broadly divided into a thread that studies the theoretical properties of algorithms for estimation or prediction in a regression mixture model as well as another thread that studies the empirical and theoretical properties of transformers on learning regression models. Due to space considerations, we present a more detailed overview in Appendix~\ref{sec:rwork}.

\subsection{Generative model for data}
\label{sec:distribution-of-prompts}
Underlying the mixture of linear regressions, we consider the discrete 
mixture
\begin{equation}\label{eqn:mixture-dist}
\mixturedist \defn \frac{1}{\numcomponents} \sum_{i=1}^\numcomponents \delta_{w_i^\star}, 
\end{equation}
where $\{w_i^\star\}_{i=1}^\numcomponents \in \R^\dimension$ are normalized such that $\twonorm{w_i^\star} = \sqrt{\dimension}$ for each $i \in [\numcomponents]$. We consider \emph{prompts} or \emph{batches}, denoted 
$P = (x_1, y_1,\dots, x_{\promptlen}, y_{\promptlen}, x_{\promptlen + 1})$. 
Here, for noise level $\sigma \geq 0$, we have
\begin{equation}\label{eqn:generative-model}
w \sim \mixturedist, \quad 
x_{i} \simiid \Normal{0}{I_\dimension}, 
\quad \mbox{and} \quad 
y_i \mid x_i \sim \Normal{\langle w, x_i\rangle}{\sigma^2}.\footnote{When $\sigma = 0$, 
by $\Normal{v}{0}$ we mean the point mass $\delta_v$.}
\end{equation}
The goal is then to predict $y_{k+1}$, the label for the query $x_{k+1}$.

\subsection{Transformers}\label{sec:transformers}

Transformers are deep neural networks that map sequences 
to sequences~\citep{VasEtAl2017}. In this work, we specifically 
focus on decoder-only, \emph{autoregressive transformers}. These models 
are comprised of multiple layers that map an input matrix 
$H \in \R^{p \times q}$ to an output matrix $H' \in \R^{p \times q}$. 
Here $p$ denotes the hidden dimension, and $q$ is corresponds to the
number of input tokens.
The output is then fed successively to more such layers. 
Since the computation in each layer is the same (apart from parameters), we describe the computation occurring in a single layer. 
Write $h_j = (H_{ij})_{1 \leq i \leq p}$ for the $j$th column of 
$H$, and $h_j'$ for the $j$th column of $H'$. Additionally, 
the prefix matrix $H_{:i}$ is the $p \times (i-1)$ submatrix 
of $H$ obtained by concatenating the first $i-1$ columns of $H$.\footnote{In 
the case $i = 1$, the submatrix can be interpreted as $0$.} 

A layer is parameterized by a sequence of weights. 
Let $\numheads$ denote the number of attention heads and 
$\dimattention$ denote the hidden dimension for the attention layer 
and $\dimff$ denote the hidden dimension for the feedforward (\ie dense) layer.
Then, a layer has the following weights: 
\begin{equation}
\label{eqn:transformer-params}
\begin{gathered}
\{\QueryMat_i\}_{i=1}^\numheads, 
\{\ValueMat_i\}_{i=1}^\numheads, 
\{\KeyMat_i\}_{i=1}^\numheads, 
\subset \R^{\dimattention \times p}, 
\\
\{\CombineMat_i\}_{i=1}^\numheads \subset \R^{p \times \dimattention}
\quad 
\FFInMat \in \R^{\dimff \times p}, 
\quad \mbox{and} \quad
\FFOutMat \in \R^{p \times \dimff}
\end{gathered}
\end{equation}
for each column $i \in [p]$, the computation proceeds in the following 
fashion.

\paragraph{Self-attention:}
The layer begins with computing the attention vector, $a_i \in \R^p$, 
by
\begin{align*}
s_{ij} &\defn \softmax\Big(\big(\KeyMat_j H_{:i} \big)^\T \QueryMat_j h_i\Big), 
\quad \mbox{for}~j \in [\numheads], \quad \mbox{and,}\\
a_i &\defn \sum_{j=1}^\numheads 
\CombineMat_j  \ValueMat_j H_{:i} s_{ij}
\end{align*}
Above, with a slight abuse of notation, we define 
for any integer $\ell > 0$, 
$\softmax \colon \R^\ell \to \R^\ell$ by the formula 
$\softmax(v) = (\e^{v_t}/\sum_{t'=1}^\ell \e^{v_{t'}})_{t=1}^{\ell}$.
Note that, above, $s_{ij} \in \R^{i-1}_+$.\footnote{When 
$i = 1$, $s_{ij} = 0$.}

\paragraph{Feedforward network:} 
The layer then continues by passing the attention vector (along with
the original input column $h_i$) through a nonlinear dense layer. This is 
defined by
\[
h'_i \defn a_i + h_i + \FFOutMat \nonlin_\bullet( \FFInMat \layernorm(a_i + h_i)) 
\]
Above the notation $\nonlin_\bullet$ indicates that the map $\nonlin
\colon \R \to \R$ is 
applied \emph{componentwise} to its argument. In this work 
we take the nonlinearity to be the Gaussian error linear unit (GeLU)~\citep{HenGim16}
which is defined by
\[
\sigma(u) = \frac{u}{2} \Big(1 + \mathrm{erf}\Big(\frac{u}{\sqrt{2}}\Big)
\Big), \quad \mbox{for any}~u\in \R.
\]
Above, $\mathrm{erf}$ denotes the Gauss error function. 
The function $\layernorm \colon \R^p \to \R^p$ denotes layer normalization~\citep{BaEtAl16}, 
and is given by 
\[
\layernorm(v) = \sqrt{p} \frac{v - \overline v \1_p}{
\twonorm{v - \overline v\1_p}}, 
\quad \mbox{where} \quad \overline{v} = \frac{1}{p} \sum_{i=1}^p v_i.
\]
This is a form of standardization where $\overline v$ is 
interpreted as the mean (averaging the components) and 
$\|v - \overline v\1_p\|_2^2/p$ is interpreted as the variance 
(averaging the squared deviation to the mean). 

\section{Representation}

In this section, we prove that transformers can actually represent the minimum mean squared error procedure. Indeed, let $f \colon P \mapsto \hat y \in \R$, by any procedure which takes a prompt $P$ and outputs an estimate $\hat y$ on the query, and define the mean squared error (MSE) by
\[
\MSE(f) \defn \E_{P} \Big[(f(P) - y_{\promptlen + 1})^2\Big].
\]
Then by standard Bayesian decision theory, under the observational model described in Section~\ref{sec:distribution-of-prompts}, it follows that 
the mean squared error is minimized at the posterior mean $f^\star_\pi$, 
which is given by
\begin{equation}\label{eqn:posterior-mean-alg}
f^\star_\pi(P) = \langle \hat w(P), x_{\promptlen + 1}\rangle \quad \mbox{where}\quad
\hat w(P) \defn \frac{\sum_{j=1}^\numcomponents w_j^\star
\exp\Big(-\frac{1}{2 \sigma^2} \sum_{i=1}^\promptlen 
(\langle w_j^\star, x_i \rangle - y_i)^2 \Big)}{\sum_{\ell=1}^\numcomponents 
\exp\Big(-\frac{1}{2 \sigma^2} \sum_{i=1}^\promptlen 
(\langle w_\ell^\star, x_i \rangle - y_i)^2 \Big)}.
\end{equation}
Formally, $\MSE(f) \geq \MSE(f^\star_\pi)$, for all (measurable) $f$. Note 
above that $\hat w$ does not depend on $x_{k+1}$. 

Then our main result is that the function $f_\pi^\star$ can be 
computed by a transformer. 
\btheo
\label{thm:transformers-can-represent-pm}
There is an autoregressive transformer which implements the function $f_\pi^\star$
as defined in equation~\eqref{eqn:posterior-mean-alg}.
\etheo 
\noindent See Section~\ref{sec:proof-of-transformers-representation} for a 
proof of this claim. 

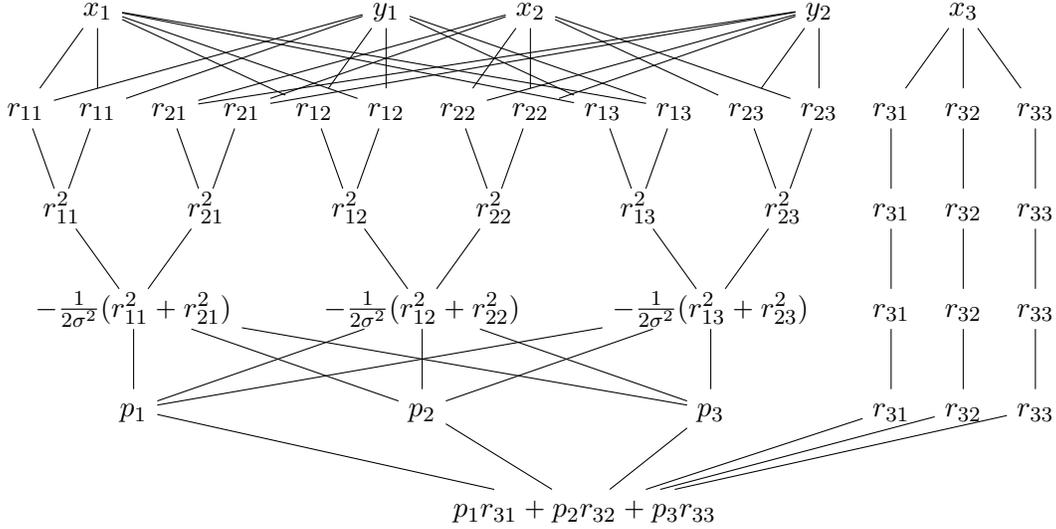
\begin{figure}
    \centering
    {\begin{forest}
for tree = {
    l sep = 8mm,
    s sep = 2mm,
    grow=north,
    text height=1.4ex, text depth=0.2ex,     math content
            }
[{p_1 r_{31} + p_2 r_{32} + p_3 r_{33}}
    [r_{33} [r_{33} [r_{33} [r_{33}, name=res33 ]]]]
    [r_{32} [r_{32} [r_{32} [r_{32} [x_3, name=x3]]]]]
    [r_{31} [r_{31} [r_{31} [r_{31}, name=res31 ]]]]
    [ {p_3}, name=prob3,
            [{-\tfrac{1}{2\sigma^2}(r_{13}^2 + r_{23}^2)}, name=sum3,
                [{r_{23}^2}, 
                    [r_{23}, name=res23c1,  
                        [y_2, name=y2]]
                    [r_{23}, name=res23c2 ]
                ]
                [{r_{13}^2}, 
                    [r_{13}, name=res13c1 ]
                    [r_{13}, name=res13c2 ] 
                ]   
            ]
    ]
          [{p_2}, name=prob2,
        [{-\tfrac{1}{2\sigma^2}(r_{12}^2 + r_{22}^2)}, name=sum2, 
            [{r_{22}^2}, 
                [r_{22}, name=res22c1,
                [x_2, name=x2]
                ]
                [r_{22}, name=res22c2 ] 
            ]
            [{r_{12}^2} , 
                [r_{12}, name=res12c1,
                    [y_1, name=y1]
                ]
                [r_{12}, name=res12c2 ]  
            ]
        ]
    ]
    [{p_1}, name=prob1,
        [{-\tfrac{1}{2\sigma^2}(r_{11}^2 + r_{21}^2)}, name=sum1, 
            [{r_{21}^2}, 
                [r_{21}, name=res21c1]
                [r_{21}, name=res21c2 ] 
            ]
            [{r_{11}^2}, 
                [r_{11}, name=res11c1,
                [x_1, name=x1]
                ]
                [r_{11}, name=res11c2 ]
            ]
        ], 
    ]
]
\draw (sum2) -- (prob1);
\draw (sum2) -- (prob3);
\draw (sum3) -- (prob1);
\draw (sum3) -- (prob2);
\draw (sum1) -- (prob2);
\draw (sum1) -- (prob3);
\draw (x3) -- (res31);
\draw (x3) -- (res33);
\draw (x1) -- (res11c2);
\draw (x1) -- (res12c1);
\draw (x1) -- (res12c2);
\draw (x1) -- (res13c1);
\draw (x1) -- (res13c2);
\draw (y1) -- (res11c2);
\draw (y1) -- (res11c1);
\draw (y1) -- (res12c2);
\draw (y1) -- (res13c1);
\draw (y1) -- (res13c2);
\draw (x2) -- (res21c2);
\draw (x2) -- (res21c1);
\draw (x2) -- (res22c2);
\draw (x2) -- (res23c1);
\draw (x2) -- (res23c2);
\draw (y2) -- (res21c2);
\draw (y2) -- (res21c1);
\draw (y2) -- (res22c2);
\draw (y2) -- (res22c1);
\draw (y2) -- (res23c2);
\end{forest}}
    \caption{Illustration of an arithmetic circuit, implementable by a transformer, that computes the posterior mean as defined in display~\eqref{eqn:posterior-mean-alg}. 
    Here, $r_{ij}$ are residuals as defined in 
    display~\eqref{defn:residuals} and $p_j$ are probabilities 
    obtained via a softmax operation, as defined in display~\eqref{defn:posterior-probs}. See main text for a description of the computation occurring at each level. }
    \label{fig:example-circuit}
\end{figure}

For an illustration of the underlying idea behind Theorem~\ref{thm:transformers-can-represent-pm}, see Figure~\ref{fig:example-circuit} for an arithmetic circuit that computes the function $f_\pi^\star$, in the case $\numcomponents = 3, \promptlen=2$.
The objects $r_{ij}$ are residuals, defined as 
\begin{equation}\label{defn:residuals}
r_{ij} = \langle w_j^\star, x_i \rangle - y_i \1\{i \neq k+1\}, 
\quad \mbox{for}~i\in[\promptlen],~j\in[\numcomponents].
\end{equation}
The first layer computes the values $\{r_{ij}\}$, the second layer 
computes the squares of these values, the third layer computes the (scaled) sum 
of these values over the index $i$, which runs over the samples in the prompt, excluding the query. The fourth layer, computes the softmax of these sums, 
\begin{equation}\label{defn:posterior-probs}
p_j \defn
\frac{
\exp\Big(-\frac{1}{2 \sigma^2} \sum_{i=1}^\promptlen 
(\langle w_j^\star, x_i \rangle - y_i)^2 \Big)}{\sum_{\ell=1}^\numcomponents 
\exp\Big(-\frac{1}{2 \sigma^2} \sum_{i=1}^\promptlen 
(\langle w_\ell^\star, x_i \rangle - y_i)^2 \Big)}, 
\quad \mbox{for}~j\in [\numcomponents]
\end{equation}
And the final layer computes 
\[
\sum_{j=1}^\numcomponents p_j r_{(k+1), j} = 
\Big\langle \sum_{j=1}^\numcomponents p_j w_j^\star, x_{k+1} \Big\rangle 
= \langle \hat w(P), x_{k+1} \rangle = f^\star_\pi(P)
\]
where the last equation follows from the definitions in display~\eqref{eqn:posterior-mean-alg}. 
Therefore, the circuit depicted in Figure~\ref{fig:example-circuit} is able to compute 
the posterior mean $f_\pi^\star$, at least for the choices $\promptlen = 2, \numcomponents=3$. Generalizing the circuit to general $(\promptlen, \numcomponents)$ is straightforward; therefore, our proof amounts to exploiting the circuit and demonstrating that each operation: linear transforms in the first and final layers, squaring in the second layer, summation in the third layer, softmax in the fourth layers are all implementable by a transformer. 

\section{Experimental results}

In this section, we present results of training transformers on batches as described in Section~\ref{sec:distribution-of-prompts}. Our methodology closely follows the training procedure described in~\citep{garg2022can}. In the notation of Section~\ref{sec:transformers}, our transformer models set the hidden dimension as $p = 256$, feedforward network dimension as $\dimff = 4p = 1024$, and the number of attention heads as $\numheads = 8$. Our models have 12 layers. Additional details on the training methodology can be found in Appendix~\ref{sec:training-methodology}. We also release our training and simulation code along with this paper. 

\subsection{Transformers can learn mixtures of linear regressions}
To begin with, we investigate the performance of transformers on 
mixture models with various numbers of components and varying noise 
levels. We plot the performance of the transformer when prompted 
with a prompt $P$ of length $\promptlen$, for $1\leq \promptlen \leq 60$. 
The normalized MSE is the mean-squared error between the true 
labels and the estimated labels, divided by the dimension $\dimension = 20$.
\begin{figure}[h!t]
\includegraphics[width=0.49\linewidth]{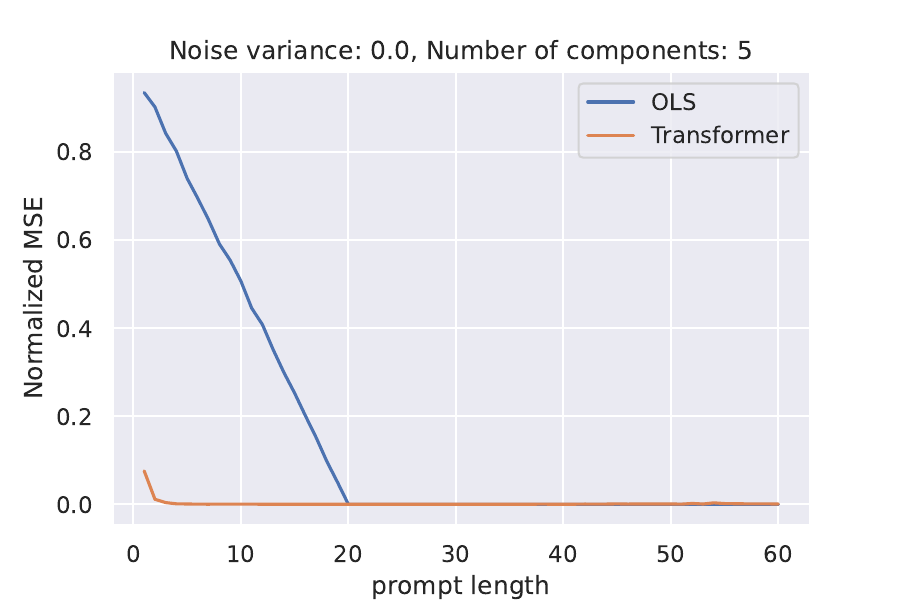}
\includegraphics[width=0.49\linewidth]{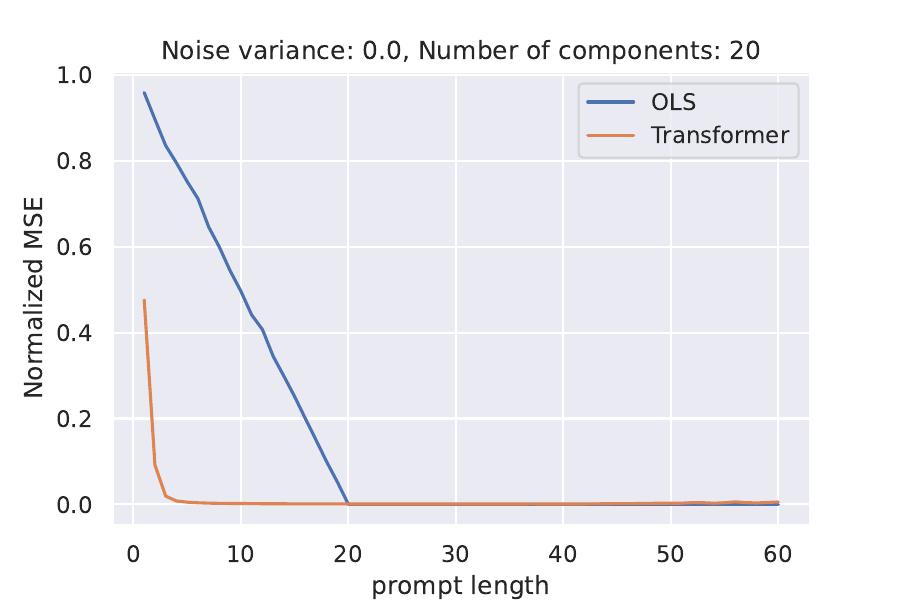}
\includegraphics[width=0.49\linewidth]{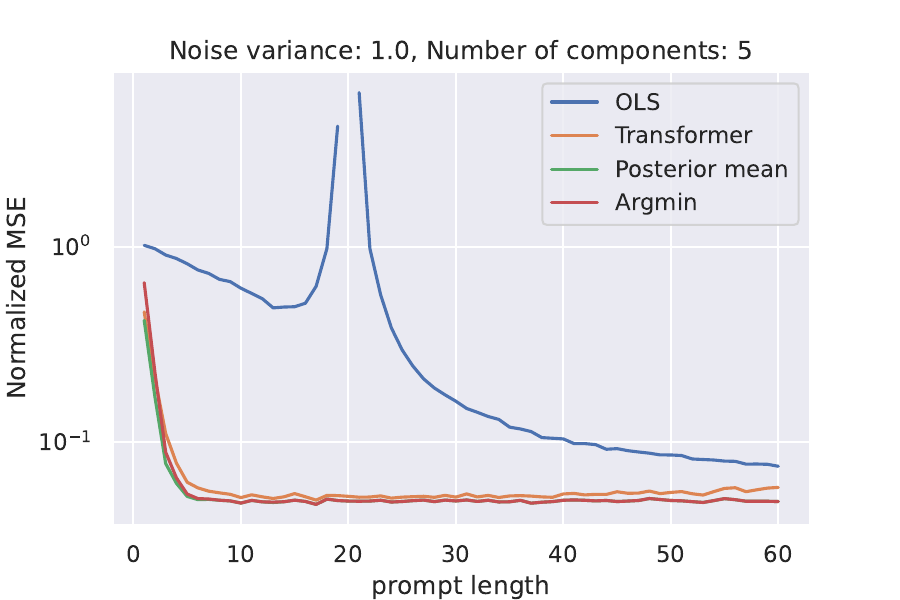}
\includegraphics[width=0.49\linewidth]{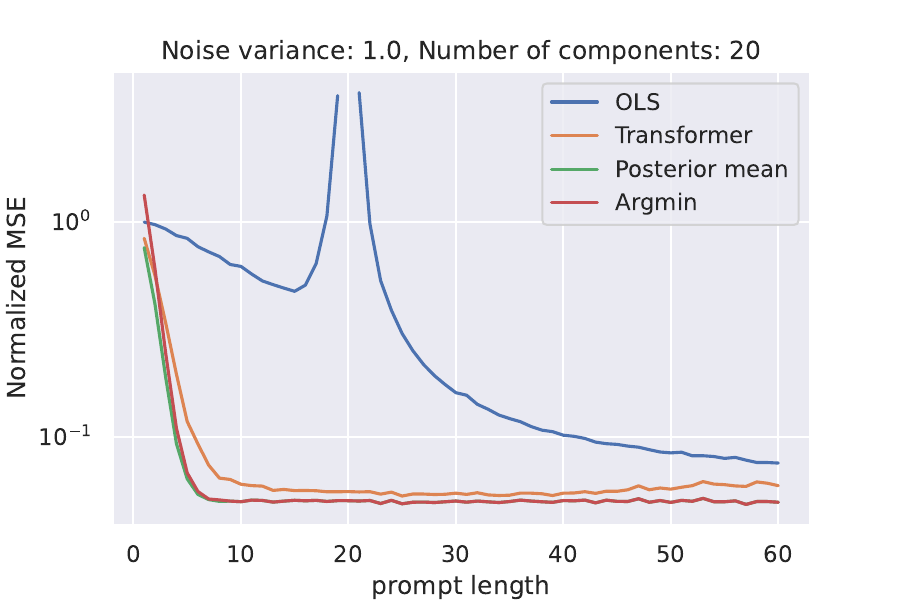}
\caption{Transformer model trained on mixture of linear regressions data with $5$ and $20$ components. Top row is mixture data with no noise added to $y$ ($\sigma = 0$) and bottom row is mixture data with noise added $(\sigma = 1)$.}
\end{figure}
Above, the algorithms that we compare against are:\footnote{For interpretability of the figures, we omit the oracle algorithms above in the noiseless case ($\sigma = 0$) as 
the error is multiple orders of magnitude smaller than the data-driven procedures.}
\begin{itemize}
    \item \emph{Ordinary least squares (OLS).} 
    For a prompt of length $\promptlen$, computes an element $\hat w \in \argmin_{w \in \R^\dimension} 
    \sum_{j \leq \promptlen} (w^\T x_j - y_j)^2.$
    Outputs $\hat y_{\promptlen + 1} = \hat w ^\T x_{\promptlen + 1}$.
    \item \emph{Posterior mean.} 
    This is an oracle algorithm. Given a prompt $P$ of length 
    $\promptlen$, computes the posterior mean $\hat y_{k+1} = f^\star_\pi(P)$, 
    as defined in display~\eqref{eqn:posterior-mean-alg}.
    \item \emph{Argmin.} 
    This is an oracle algorithm. Given a prompt $P$ of length $\promptlen$, 
    computes 
    \begin{equation}\label{eqn:argmin-alg}
    \hat w = \argmin_{w \in \{w^\star_j\}_{j=1}^\numcomponents} 
    \sum_{j \leq \promptlen}
     \sum_{j \leq \promptlen} (w^\T x_j - y_j)^2.
    \end{equation}
    The prediction is then $\hat y_{\promptlen + 1} =  \hat w ^\T x_{\promptlen + 1}$
\end{itemize}
Strikingly, we see that the transformer prodictions are as good as---or nearly as good as---the oracle procedures which have knowledge of the true 
mixture components $\{w_j^\star\}$. It is important to note that OLS is suboptimal in general for mixtures of linear regressions. Nonetheless, the transformer is performing much better than OLS, indicating the trained transformer implements a better predictor to  adapted to the mixtures of linear regressions setting. 

\subsection{Comparison of performance for fixed training set size}

Next, we investigate whether or not transformers learn mixtures of linear regressions in a sample efficient way. To do this, we depart slightly from the training methodology in~\citep{garg2022can}. We first sample a fixed training set of size $n \in \{15000, 30000, 45000, 60000\}$. Then---with some hyperparameter tuning to avoid overfitting, as well as a modification to the curriculum training, described in Appendix~\ref{sec:fixed-sample-size-training}---we train the transformer as in that paper. We then compare the inference time performance by computing the mean-squared error on prompts of length $\promptlen \in [1, 60]$. 

\begin{figure}[h!t]
\includegraphics[width=0.49\linewidth]{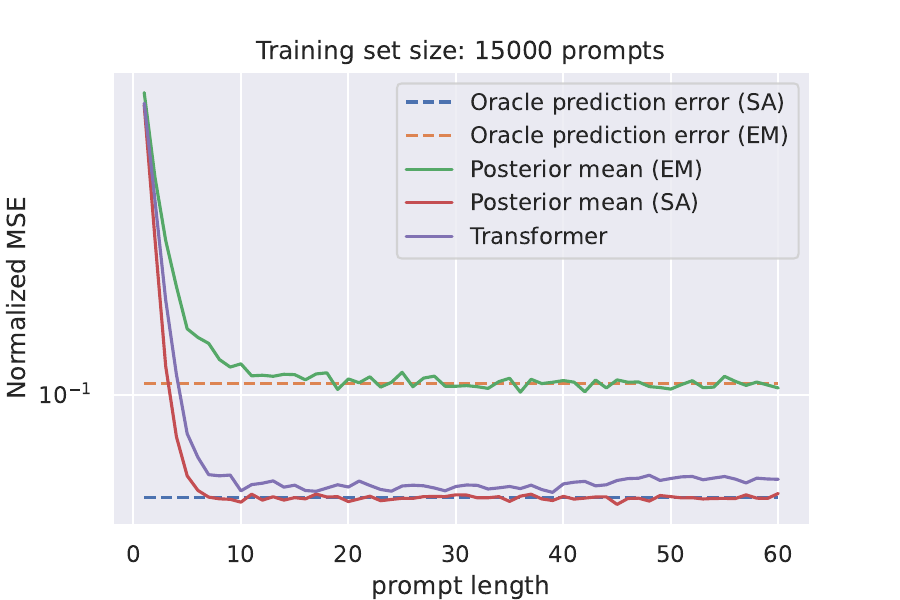}
\includegraphics[width=0.49\linewidth]{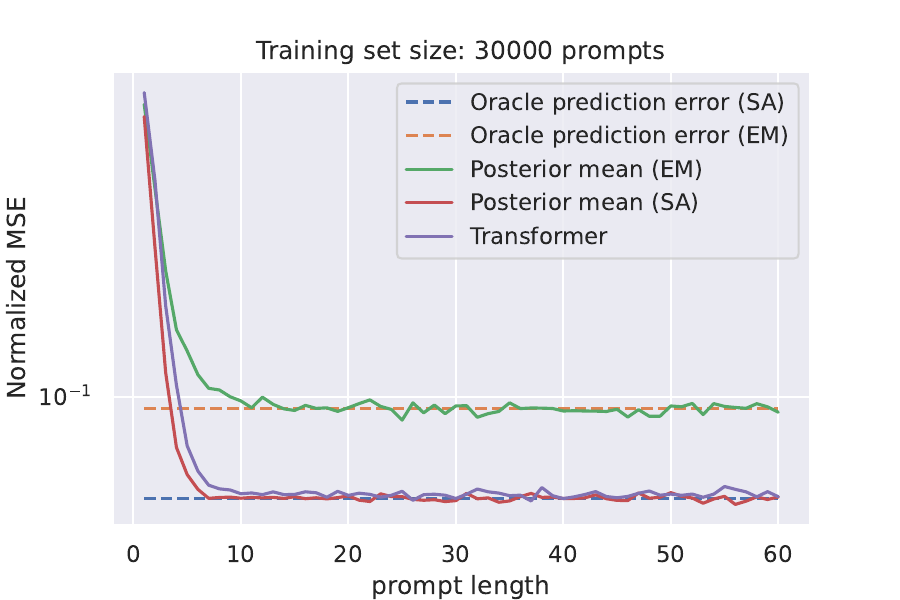} \\ 
\includegraphics[width=0.49\linewidth]{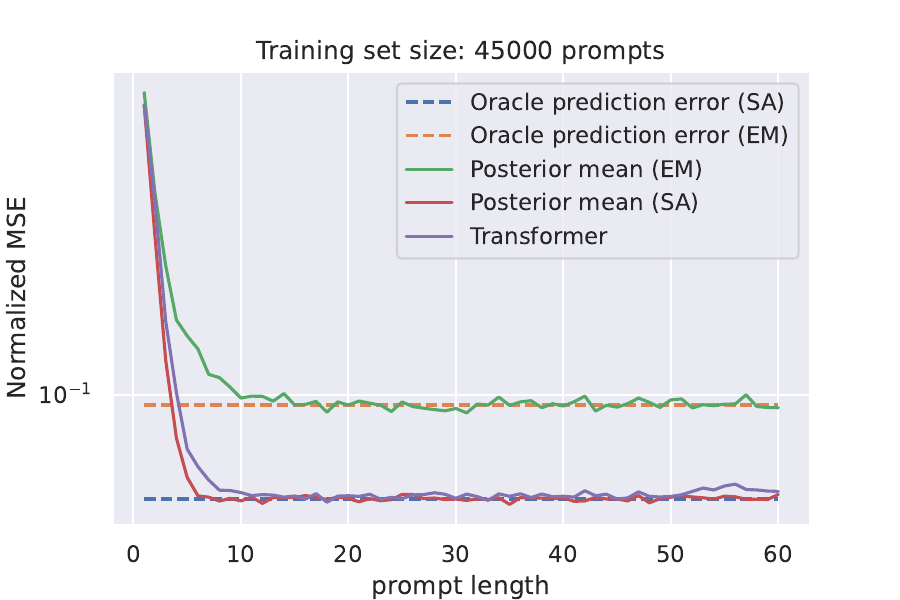}
\includegraphics[width=0.49\linewidth]{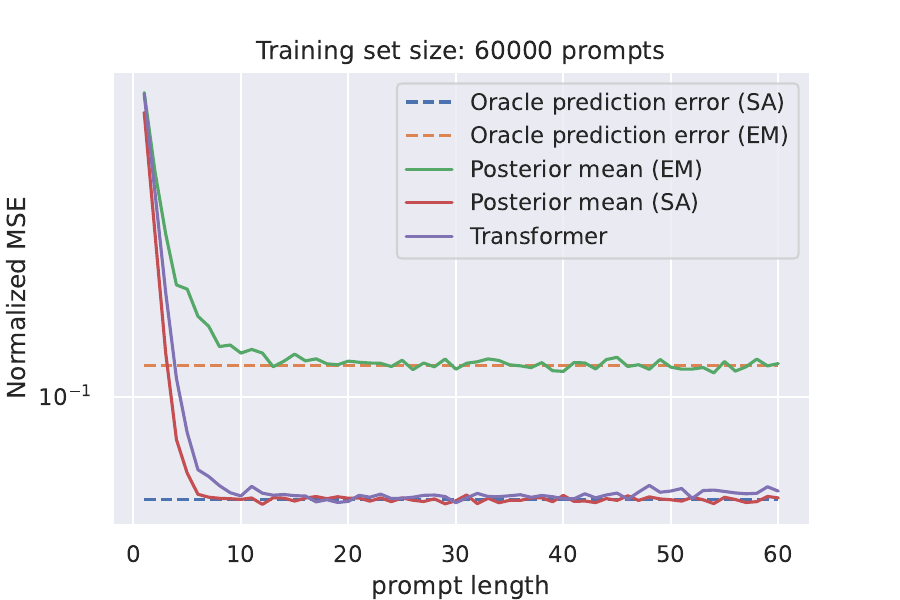} 
\caption{Comparison of EM, to subspace algorithm (SA) in~\citep{jain2023linear}, and 
Transformers.}
\label{fig:sample-size}
\end{figure}

The results of our simulation are shown below in Figure~\ref{fig:sample-size}
We compared against two other procedures, which have the form of ``plug-in'' procedures: 
\begin{itemize}
    \item \emph{Posterior mean, EM weights.}
    Here, we first estimate the component means $w_j^\star$ using batch 
    expectation-maximiation (see Appendix~\ref{sec:EM} and 
    Algorithm~\ref{alg:EM} therein for details). Then, 
    we form $\hat \pi$, the uniform distribution over the estimated weights, 
    and then predict $y_{\promptlen+1}$ by $f^\star_{\hat \pi}(P)$.
     \item \emph{Posterior mean, SA weights.}
     We follow the approach above, but  estimate the weights by using 
     the subspace algorithm (SA), which is Alg.~1 in~\citep{jain2023linear}.
\end{itemize}

Note that the `oracle prediction error' quantities appearing in 
Figure~\ref{fig:sample-size} are essentially the best possible error 
achievable using the weights estimated by the set of 
weights $\widehat \cW$ estimated by an algorithm. 
Before normalization by the dimension, it is the noise level plus 
\[
\frac{1}{\numcomponents} 
\sum_{j=1}^\numcomponents \min_{w \in \widehat \cW} \|w_j^\star - w\|_2^2, 
\]
which is easily verified to be the prediction error with oracle knowledge 
of the nearest element in $\hat \cW$ to the true component mean $w_j^\star$, under our observational model~\eqref{eqn:generative-model}.
The main take-away from this simulation is that the transformer is able to get very close to the performance of the state-of-the-art model-specific algorithms, even when keeping the sample size the same.  

\subsection{What is the transformer actually learning?}

\begin{figure}[h!t]
\includegraphics[width=0.49\linewidth]{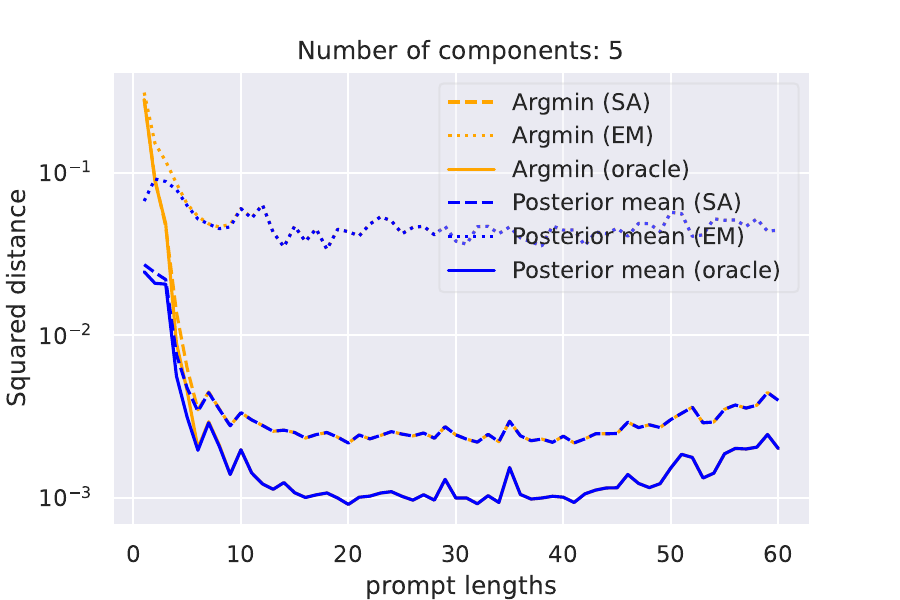}
\includegraphics[width=0.49\linewidth]{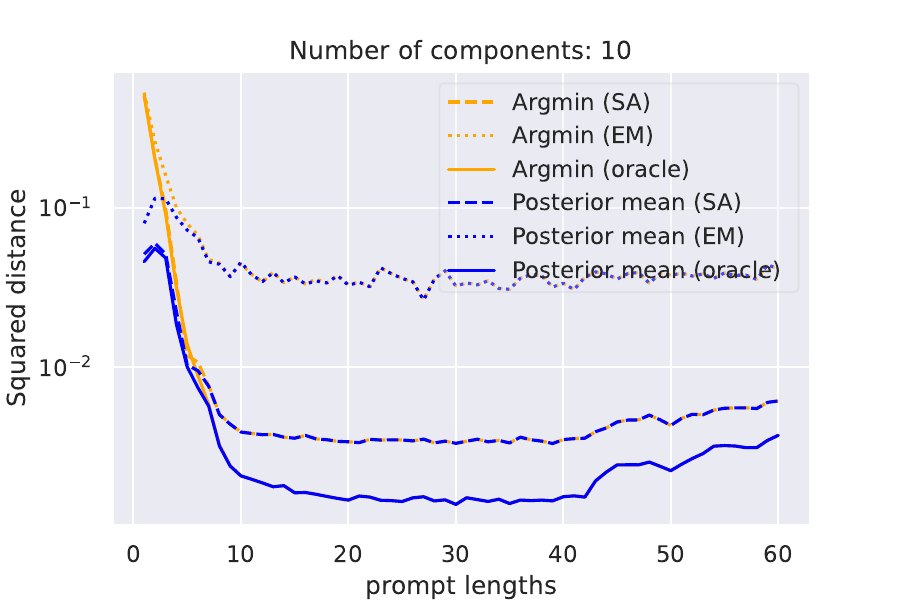}\\
\includegraphics[width=0.49\linewidth]{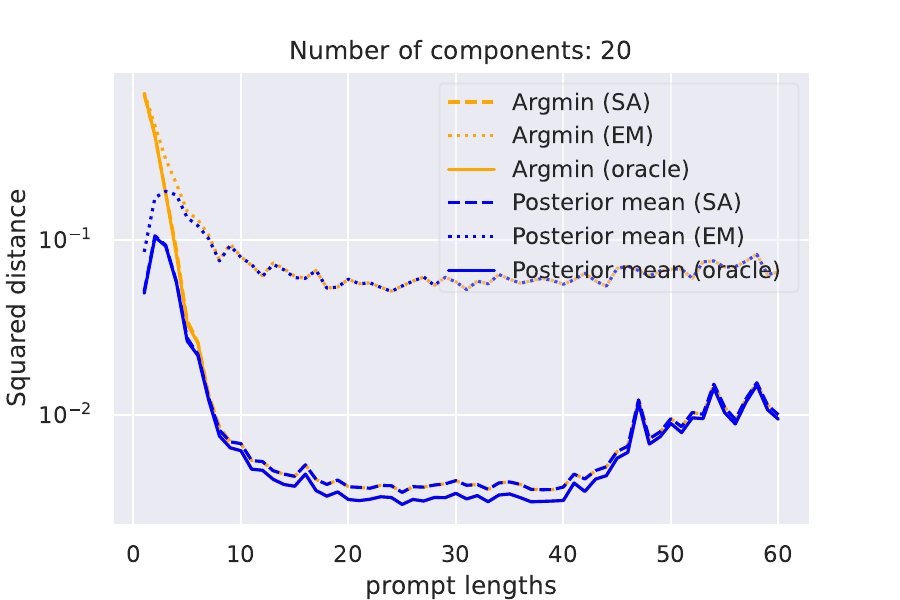}
\includegraphics[width=0.49\linewidth]{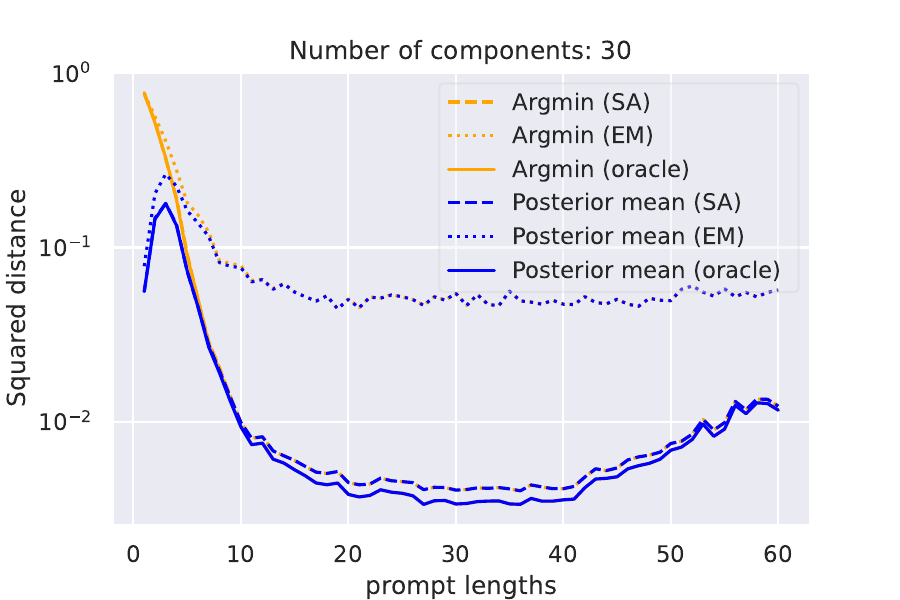}
\caption{Comparing inference-time predictions from transformer versus 
posterior mean and argmin procedures with oracle or estimated weights. In these simulations, the noise level is set as $\sigma = 1.0$.}
\label{fig:nearest-alg}
\end{figure}

In this section, we try to understand somewhat better, what algorithm the transformer
is implementing at inference time. To do this, we define the squared distance, for two algorithms 
$f, g$ that map a prompt $P$ of length $k$ to a prediction $\hat y_{\promptlen+1}$ of $x_{\promptlen+1}$:
\[
d_k^{\rm sq}(f, g) \defn \E_{P} \Big[(f(P) - g(P))^2\Big], 
\quad \mbox{where}~k\geq 1.
\]
Figure~\ref{fig:nearest-alg} depicts $k$ versus $d^{\rm sq}_k(f, g)$, taking $f$ to be the transformer, and $g$ to be a candidate algorithm listed below, as 
$k$ varies between $1$ and $60$.
The compared algorithms are:
\begin{itemize}\itemsep-0.1em
    \item \emph{Posterior mean, oracle weights.} Outputs $f^\star_\pi(P)$ on prompt $P$. 
    \item \emph{Posterior mean, SA weights.}
    Outputs $f^\star_{\hat \pi}(P)$, with $\hat \pi$ from the subspace algorithm (SA). 
    \item \emph{Posterior mean, EM weights.} 
    Same as above, but $\hat \pi$ from expectation-maximization (EM).
    \item \emph{Argmin, oracle weights.} Outputs $\hat w(P)^\T x_{\promptlen + 1}$ 
    where $\hat w(P)$ follows display~\eqref{eqn:argmin-alg}.
    \item \emph{Argmin, SA weights.} Outputs $\hat w(P)^\T x_{\promptlen + 1}$
    where $\hat w(P)$ follows display~\eqref{eqn:argmin-alg}, with 
    $w_j^\star$ replaced by SA-estimated weights. 
    \item \emph{Argmin, EM weights.} Same as above, but with EM-estimated weights. 
\end{itemize}

As seen from Figure~\ref{fig:nearest-alg}, 
in all of the simulated settings, 
the algorithm closest to the transformer at inference time is the posterior mean procedure, with the oracle weights. Impressively, this observation holds regardless of our choice of the number of mixture components. 

\subsection{Evaluation on covariate shift and label shift}
\label{sec:distribution-shift}
\begin{figure}[h!t]
\includegraphics[width=0.49\linewidth]{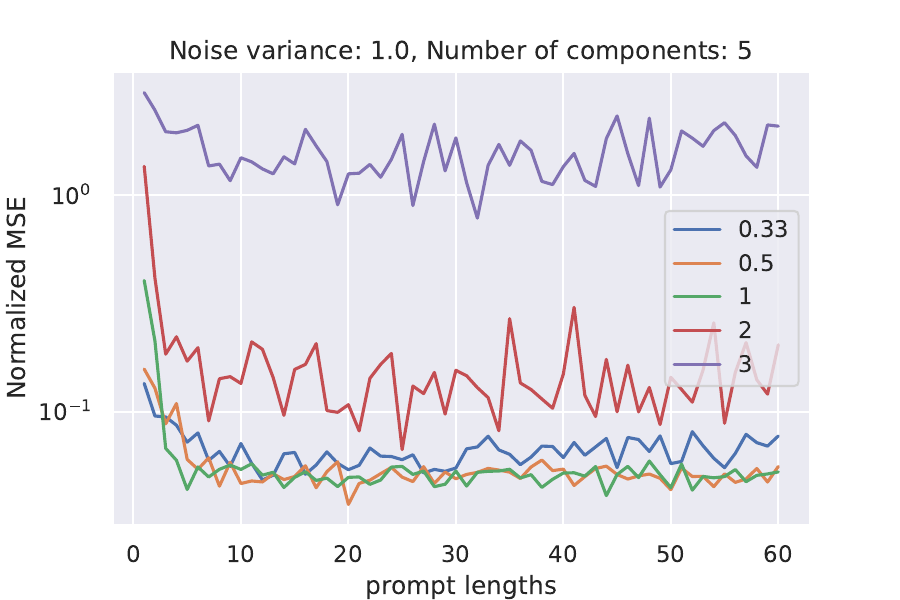}
\includegraphics[width=0.49\linewidth]{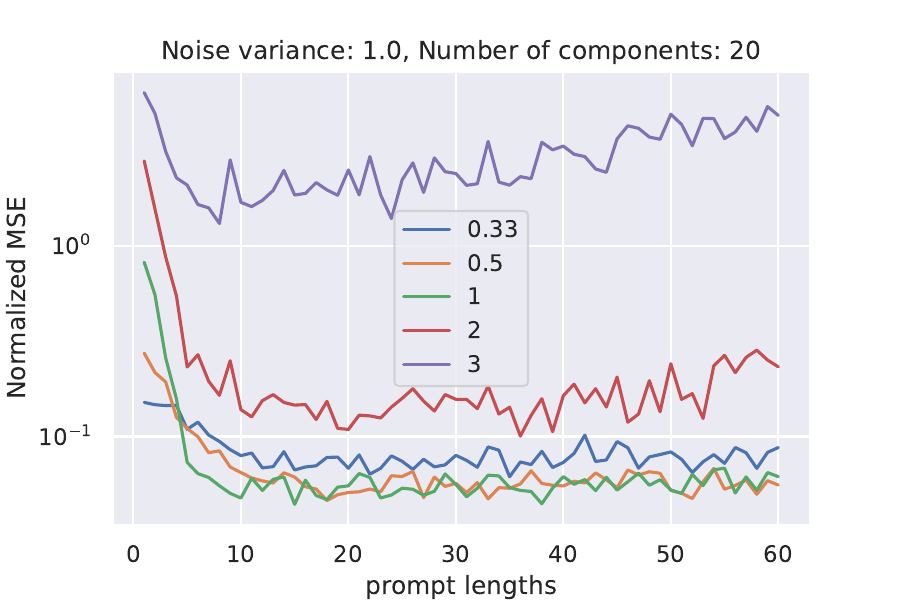}
\caption{Evaluating transformer performance on covariate shifts.}
\label{fig:covariate-shift}
\end{figure}

\begin{figure}[h!t]
\includegraphics[width=0.49\linewidth]{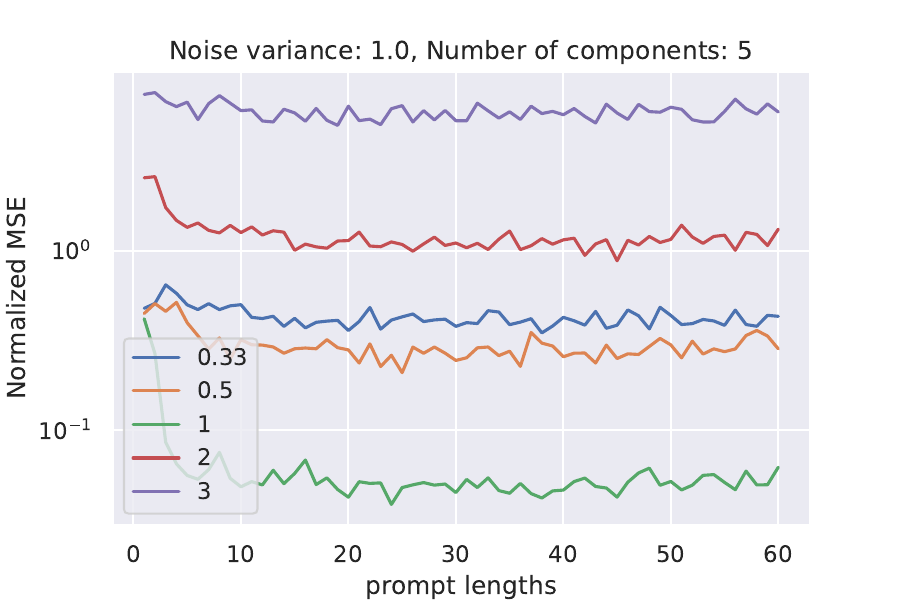}
\includegraphics[width=0.49\linewidth]{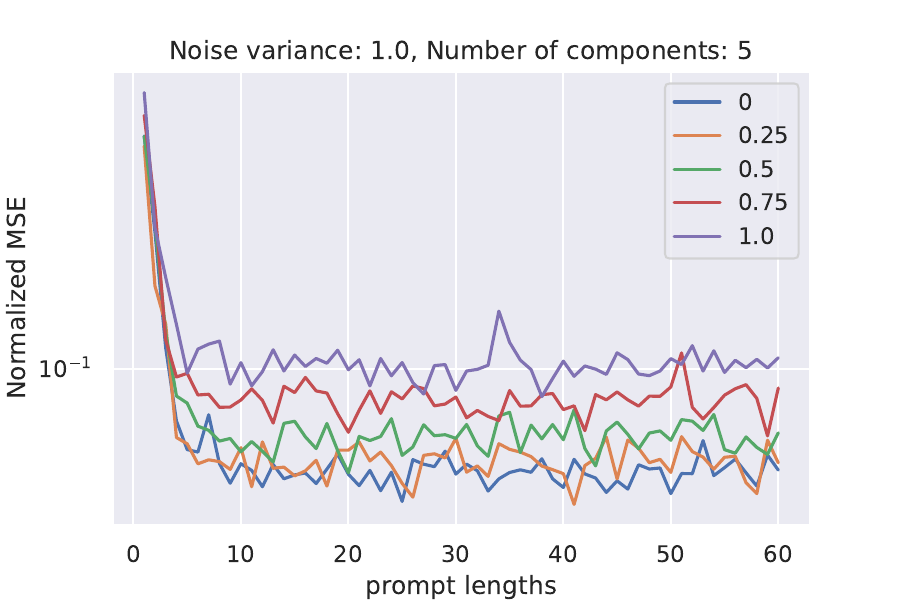}
\\
\includegraphics[width=0.49\linewidth]{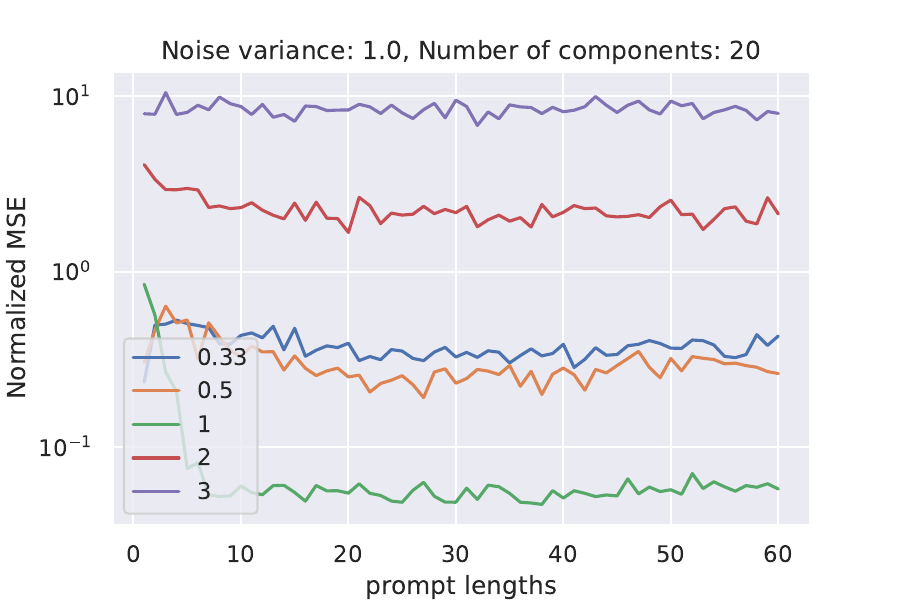}
\includegraphics[width=0.49\linewidth]{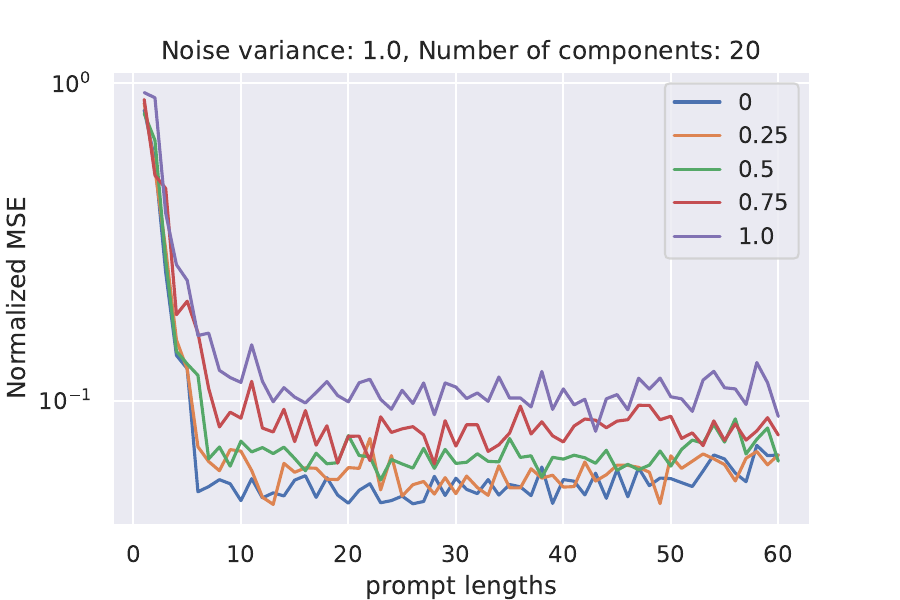}
\caption{Evaluating transformers on weight scaling (left panels) and weight shift (right panels).}
\label{fig:posterior-shift}
\end{figure}

In this section, we evaluate transformers on distribution shift settings. 
The experimental results are presented in Figures~\ref{fig:covariate-shift}
and~\ref{fig:posterior-shift}. The distribution shift settings are described below, where we studied one setting of covariate shift and two settings of label shift.

\paragraph{Covariate scaling:} We evaluate the transformer on prompts of length 
$\promptlen$ where the covariates (including the query) are sampled as $x_i \sim \Normal{0}{\kappa^2 I_\dimension}$ for $i \in [\promptlen + 1]$. 
This is a shift from the training distribution when $\kappa \neq 1$. 
Figure~\ref{fig:covariate-shift} shows the results when taking $\kappa \in \{0.33, 0.5, 1, 2, 3\}$. As we see from the figure, the transformer is able to handle, 
to some extent, small shifts, such as $\kappa \in \{0.33, 0.5, 2\}$, but not shifts much larger than this. 

\paragraph{Weight scaling:} We evaluate the transformer on prompts sampled 
from the mixture distribution 
\[
\pi^{\rm scale}_\alpha \defn \frac{1}{\numcomponents} \sum_{i=1}^\numcomponents 
\delta_{\alpha w_j^\star}, 
\quad \mbox{where}~\alpha > 0.
\]
So, the weights $w_j^\star$ are scaled up or down by the factor $\alpha$. 
Note that $\pi^{\rm scale}_1 = \pi$, meaning that $\alpha = 1$ is no shift. 
The left panels of Figure~\ref{fig:posterior-shift} depict results for $\alpha \in \{0.33, 0.5, 1.0, 2, 3\}$.

\paragraph{Weight shift:} We evaluate the transformer on weights sampled from the 
mixture distribution 
\[
\pi^{\rm add}_{\varepsilon} \defn \frac{1}{\numcomponents} \sum_{i=1}^\numcomponents 
\delta_{w_j(\varepsilon)}, 
\quad \mbox{where} 
\quad 
w_j(\varepsilon) \defn w_j^\star + \frac{\varepsilon}{\sqrt{\dimension}} \1_\dimension.
\]
Thus, $\pi^{\rm add}_{\varepsilon}$ shifts each component by an additive perturbation of norm $\varepsilon$. The right panels of Figure~\ref{fig:posterior-shift} depict the results for $\varepsilon \in \{0, 0.25, 0.5, 0.75, 1.0\}$. Note that $\varepsilon = 0$ is no shift: 
$\pi^{\rm add}_{0} = \pi$. 

As seen from above, the transformer is fairly sensitive to weight scaling, as 
seen from the left panels in Figure~\ref{fig:posterior-shift}. On the other hand, 
the transformer can handle small additive weight shifts, such as $\eps = 0.25$, 
as depicted in the right panels in Figure~\ref{fig:posterior-shift}.

\paragraph{Comparison to posterior mean procedure:} 
In Appendix~\ref{sec:comparison-to-pma-on-shfit}, 
we replicate the figures above, with the change that 
in place of the transformer, we evaluate the performance of the 
posterior mean procedure, $f^\star_\pi$, defined in 
display~\eqref{eqn:posterior-mean-alg}. 
At a high-level, the posterior mean algorithm is less 
sensitive to covariate scaling, 
but exhibits similar behavior to the transformer 
on the two label shift settings. 

\section{Discussion}

In this work, we studied the behavior of transformers on mixtures of linear regressions, and showed that they can learn these mixture models near-optimally, sample-efficiently, and somewhat robustly. 
The fact that transformers----which, importantly, are general purpose prediction 
methods---can perform well in this statistically-complex mixture setting 
could be quite useful for practical problems, where 
it may be undesirable to use methods requiring a well-specified mixture model. 

Additionally, our empirical and theoretical observations also suggest
some lines of future investigation. For instance, 
in this work we have studied mixtures of linear regressions. 
However, in practice, the regression function within each component
could potentially be \emph{nonlinear}. To what extent 
do transformers perform well in these settings? 
Additionally, it would be interesting to study the in-context problem as was done in~\cite{garg2022can}, but in the mixture setting. Here, the mixture distribution would be sampled from a distribution \emph{over} mixture models for each prompt. In general, the decision-theoretic optimal method could be more complicated to compute, as implementing the posterior mean would require computing a high-dimensional integral.
Nonetheless, is it possible to approximate the optimal method 
with a trained transformer? We view this as an intriguing direction for future work. 

\newpage 
\bibliography{references}
\bibliographystyle{abbrvnat}

\newpage
\appendix

\section{Related Work}
\label{sec:rwork}

The related work can be broadly divided into two categories: (i) theoretical works on learning mixture models and (ii) analyzing theoretically and empirically the learning abilities of transformers.

In the context of (i), there are numerous works that study the well known mixed linear regression problem with a batch size of $1$~\citep{vempala2004spectral, yi2014alternating, yi2016solving, chen2019learning, li2018learning, zhong2016mixed}. In general the problem is NP-Hard as shown in~\citep{yi2016solving}. Therefore most of the above works with the exception of~\citep{li2018learning} makes the assumption that the covariates of all the mixture components are isotropic Gaussians. However, even with this strong assumption the time-complexity of all these algorithms are at least super-polynomical in $m$ rendering them impractical.

\citet{kong2020meta} pioneered the study of the problem in batch setting where they were motivated by meta-learning multiple tasks. They showed that they can recover well separated mixture of linear models from batched data with polynomial dependence on $d, m$ and the inverse of the fraction of the smallest mixture component. However, this work still had the isotropic covariate assumption. Recent work~\citep{jain2023linear} removed this assumption and further improved the sample complexity and the length of the medium size batches that is required for learning. We compare the training sample complexity of learning using transformers with that of the latter, as well as the popular EM method~\citep{zhong2016mixed}, modified to work with batched data.

In the context of (ii), following the emergence of several hundred billion parameter large language models (LLM)'s like~\citep{radford2019language, chowdhery2022palm}, it has been observed that such models can learn from few examples supplied in a prompt during inference~\citep{brown2020language}. This ability to learn in-context has been studied in simpler settings in many recent works~\citep{garg2022can, von2023transformers, MaEtAl2023, zhang2023trained}. \citep{garg2022can} showed empirically that transformers can learn to perform linear regression in context. \citep{MaEtAl2023} then showed that transformers can represent gradient decent for linear regression in context. A similar result was shown in~\citep{von2023transformers} but using linear self attention. \citet{zhang2023trained} go one-step further by showing that gradient flow in linear self-attention based transformers can learn to do population gradient decent for linear regression. More general algorithm learning behavior has been demonstrated in~\citep{li2023transformers} and they also provide stability bounds for in-context learning.

Note that none of these prior works imply ability of transformers to learn mixture models from batch or non batch setting. \citet{muller2021transformers, ahuja2023context} look at in context learning from a Bayesian perspective. \citet{muller2021transformers} show that transformers fitted on the respective prior can emulate Gaussian processes. \citep{ahuja2023context} has a section on learning multiple function classes in-context where they empirically study gaussian mixture models with two mixture components. However, they do not study the representation learning problem and training sample complexity is not investigated in depth. 

\section{Proof of Theorem~\ref{thm:transformers-can-represent-pm}}
\label{sec:proof-of-transformers-representation}

In this section, we present the proof of Theorem~\ref{thm:transformers-can-represent-pm}. 
We begin, in Section~\ref{sec:transformer operators} 
by stating some preliminaries, such as the 
necessary operators we need to show that the transformer can implement. We then present the proof, 
assuming that these operators are transformer-representable
in Section~\ref{sec:full-proof}. 
Finally, the proof of the representation capacity of 
these operators by transformers 
is provided in Section~\ref{sec:proof-prop-implementable-ops}.

\subsection{Operators that a transformer can implement}
\label{sec:transformer operators}
We now list some operators, for a matrix 
$H \in \R^{p \times q}$ that output a 
matrix $H' \in \R^{p \times q}$. The following 
list includes all the operators we need.
\begin{itemize}
\item $\copydown(H; k, k', \ell, \cI)$: \quad  
For columns with index $i \in \cI$, 
outputs $H'$ where $H'_{k':\ell', i} = H_{k:\ell, i}$, 
and the remaining entries are unchanged. Here, 
$\ell' = k' + (\ell-k)$ and $k' \geq k$, so that entries are copied ``down" within columns $i \in \cI$. Note, 
we assume $\ell \geq k$ and that $k' \leq q$ so that the 
operator is well-defined. 
\item $\copyover(H; k, k', \ell, \cI)$:\quad 
For columns with index $i \in \cI$, 
outputs $H'$ with $H'_{k':\ell', i} = H_{k:\ell,i-1}$. 
The remaining entries stay the same. Here entries from column $i- 1$ are copied ``over'' to column $i$. 
\item $\mulop(H; k, k', k'', \ell, \cI)$: \quad 
For columns with index $i \in \cI$, outputs 
$H'$ where 
\[
H'_{k'' + t, i} = H_{k + t, i} H_{k' + t, i}, 
\quad \mbox{for}~t \in \{0, \dots, \ell - k\}. 
\]
for 
$t \in [k, \ell]$. The remaining entries stay the same.
\item $\affineop(H; k, k', k'', \ell, \ell', \ell'', W, 
W', b, \cI)$: \quad 
For columns with index $i \in \cI$, outputs $H'$ 
where \[
H'_{k'':\ell'', i} = 
W H_{k:\ell, i} + W' H_{k':\ell', i} + b.
\]
Note that $\ell'' = k'' + \delta''$ where 
$W \in \R^{\delta'' \times \delta}$, 
$W' \in \R^{\delta'' \times \delta'}$ and 
$\ell =  k + \delta$, $\ell' = k' + \delta'$. 
We assume $\delta, \delta', \delta'' \geq 0$.
The remaining entries of $H$ are copied over to $H'$, unchanged.
\item $\scaledagg(H; \alpha, k, \ell, k', i, \cI )$:
\quad 
Outputs a matrix $H'$ with entries 
\[
H_{k' + t, i} = \alpha \sum_{j \in \cI} H_{k + t, j}
\quad \mbox{for}~t \in \{0, 1, \dots, \ell-k\}. 
\] 
The set $\cI$ is causal, so that $\cI \subset [i-1]$. 
The remaining entries of $H$ are copied over to $H'$, unchanged.

\item $\softmaxop(H; k, \ell, k')$:
\quad 
For the final column $q$, outputs 
a matrix $H'$ with entries
\[
H'_{k' + t, q} = \frac{\e^{H_{k + t, q}}}{
\sum_{t' = 0}^{\ell - k} \e^{H_{k + t', q}}},
\quad \mbox{for}~t \in \{0, 1, \dots, \ell-k\}.
\]
The remaining entries of $H$ are copied over to $H'$, unchanged.
\end{itemize}

The important property of the above list of operators is that can all be implemented in a single layer of a autoregressive transformer. 

\bpr \label{prop:implementable-ops}
Each of the operators $\copydown, \copyover, \mulop, \affineop, \scaledagg$, and $\softmaxop$, can be implemented by a single 
layer of an autoregressive transformer. 
\epr 
\noindent See Section~\ref{sec:proof-prop-implementable-ops} 
for a proof of this claim.

\subsection{Proof of Theorem~\ref{thm:transformers-can-represent-pm}}
\label{sec:full-proof}

In this section, we present the proof of Theorem~\ref{thm:transformers-can-represent-pm}, 
assuming Proposition~\ref{prop:implementable-ops}. 
We need to introduce a bit of notation: 
\[
\evenindices \defn \{2j : j \in [\promptlen]\} 
\quad \mbox{and} \quad 
\oddindices \defn \{2j - 1 : j \in [\promptlen + 1]\}. 
\]
Additionally, we define 
$\clustermatrix \in \R^{\numcomponents \times \dimension}$
to have rows $w_i^\star \in \R^\dimension$, which as 
we recall from~\eqref{eqn:mixture-dist}, are the 
true mixture weights. 

We begin by assuming that the input prompt $P$ is 
provided as $H^{(0)} \in \R^{
(2\dimension + 4\numcomponents + 2) \times 
(2 \promptlen + 1)}$. 
This matrix is such that the only nonzero entries are 
$H^{(0)}_{1:\dimension, 2j - 1} = x_j \in \R^\dimension$ 
for each $j \in [\promptlen+1]$.
Additionally, 
$H^{(0)}_{1, 2j} = y_j$ for each $j \in [\promptlen]$
Then, by leveraging the operators described above, 
we can see that $f^\star(P) = H^{(9)}_{2 \dimension + 4 \numcomponents + 2, 2 \promptlen + 1}$, where 
the matrix $H^{(8)}$ is constructed by the following process: 
\begin{itemize}
\item $H^{(1)} = 
\copydown(H^{(0)}; 1, \dimension+1, \dimension, 
\oddindices)$
\item $H^{(2)} = 
\copyover(H^{(1)}; \dimension+1, \dimension+1, 
2\dimension, \evenindices)$
\item $H^{(3)} = 
\copydown(H^{(2)}; 1, 2 \dimension + 1, 1, \oddindices)$
\item $H^{(4)} = 
\affineop(H^{(3)}; \dimension + 1, 2d+1, 2\dimension+2, 
2\dimension, 2d+1, 2\dimension + \numcomponents + 1, 
\clustermatrix, \1_{\dimension \times 1}, 0,  
\evenindices \cup \{2 \promptlen + 1\})$
\item $H^{(5)} = 
\mulop(H^{(4)}; 
2\dimension + 2, 
2\dimension + 2, 
2\dimension + \numcomponents + 2, 
2\dimension + \numcomponents + 1, 
\evenindices)$
\item $H^{(6)} = 
\scaledagg(H^{(5)}; -\tfrac{1}{2\sigma^2}, 
2\dimension + \numcomponents + 2, 
2\dimension + 2\numcomponents + 1, 
2\dimension + \numcomponents + 2, 
2\promptlen + 1, 
\evenindices)$
\item $H^{(7)} = 
\softmaxop(H^{(6)}; 
2\dimension + \numcomponents + 2, 
2\dimension + 2 \numcomponents + 1, 
2\dimension + 2\numcomponents + 2)$
\item $H^{(8)} = 
\mulop(H^{(7)}; 
2\dimension + 2  ,
2\dimension + 2\numcomponents + 2, 
2\dimension + 3\numcomponents + 2, 
2 \dimension + \numcomponents + 1
\{2 \promptlen + 1\})$
\item $H^{(9)} = 
\affineop(H^{(8)};
2\dimension + 3\numcomponents + 2, 
\dimension + 1, 
2\dimension + 4\numcomponents + 1, 
2\dimension + 2\numcomponents, 
\dimension + 1, 
2\dimension + 4\numcomponents + 2, 
\1_{m}, 0,  0 , \{2\promptlen + 1\})
)$
\end{itemize}
The process above is illustrated in Section~\ref{sec:illustration}. 
By Proposition~\ref{prop:implementable-ops}, 
each operation above is 
implementable by a layer of an autoregressive 
transformer. Therefore, this completes the proof. 

\subsubsection{Illustration of proof of 
Theorem~\ref{thm:transformers-can-represent-pm}}
\label{sec:illustration}
We illustrate the 
steps taken by the transformer to implement 
the softmax operation. 
To begin with, the matrix input to the transformer is modelled as below, 
in the case where $\promptlen = 2$. 
Below, $\tilde y_i = (y_1, 0, \dots, 0) \in \R^\dimension$.
Throughout we only show the nonzero entries (\ie, missing 
rows and columns are always assumed 0). 
Then, our input is
\[
H^{(0)} = \begin{bmatrix}
x_1 & \tilde y_1 & x_2 & \tilde y_2 & x_3 
\end{bmatrix}.
\]
After the $\copydown$ operation, we have
\[
H^{(1)} = \begin{bmatrix}
x_1 & \tilde y_1 & x_2 & \tilde y_2 & x_3 \\
x_1 & 0 & x_2 & 0 & x_3  
\end{bmatrix}.
\]
After the $\copyover$ operation, we have 
\[
H^{(2)} =
\begin{bmatrix}
x_1 & \tilde y_1 & x_2 & \tilde y_2 & x_3 \\
x_1 & x_1 & x_2 & x_2 & x_3  
\end{bmatrix}.
\]
After another $\copydown$ operation, we have
\[
H^{(3)} = 
\begin{bmatrix}
x_1 & \tilde y_1 & x_2 & \tilde y_2 & x_3 \\
x_1 & x_1 & x_2 & x_2 & x_3  \\ 
0 & y_1 & 0 & y_2 & 0  
\end{bmatrix}.
\]
After the $\affineop$ operation, we have, 
\[
H^{(4)} = 
\begin{bmatrix}
x_1 & \tilde y_1 & x_2 & \tilde y_2 & x_3 \\
x_1 & x_1 & x_2 & x_2 & x_3  \\ 
0 & y_1 & 0 & y_2 & 0  \\
0 & r_1  & 0 & r_2  & 
r_3 
\end{bmatrix}.
\]
Note that $r_i = \clustermatrix x_i - y_i \1$ for 
$i \neq \promptlen +1$ and otherwise $r_{\promptlen + 1} = 
\clustermatrix x_{\promptlen + 1}$. 
After the $\mulop$ operation, we obtain
\[
H^{(5)} = 
\begin{bmatrix}
x_1 & \tilde y_1 & x_2 & \tilde y_2 & x_3 \\
x_1 & x_1 & x_2 & x_2 & x_3  \\ 
0 & y_1 & 0 & y_2 & 0  \\
0 &r_1 & 0 & r_2  & 
r_3  \\
0 &r_1^{2} & 0 & r_2^2  & 
0  \\
\end{bmatrix}
\]
Above the square should be interpreted element wise on the 
vectors $r_i$. Then, after the $\scaledagg$ operation, we obtain
\[
H^{(6)} = 
\begin{bmatrix}
x_1 & \tilde y_1 & x_2 & \tilde y_2 & x_3 \\
x_1 & x_1 & x_2 & x_2 & x_3  \\ 
0 & y_1 & 0 & y_2 & 0  \\
0 &r_1 & 0 & r_2  & 
r_3  \\
0 &r_1^{2} & 0 & r_2^2  & 
-\tfrac{1}{2\sigma^2}(r_1^2 + r_2^2)  \\
\end{bmatrix}.
\]
Then, after the $\softmax$ operation, we obtain
\[
H^{(7)} = 
\begin{bmatrix}
x_1 & \tilde y_1 & x_2 & \tilde y_2 & x_3 \\
x_1 & x_1 & x_2 & x_2 & x_3  \\ 
0 & y_1 & 0 & y_2 & 0  \\
0 &r_1 & 0 & r_2  & 
r_3  \\
0 &r_1^{2} & 0 & r_2^2  & 
-\tfrac{1}{2\sigma^2}(r_1^2 + r_2^2)  \\ 
0 & 0 & 0 & 0 & p \\ 
\end{bmatrix}.
\] 
Here, $p = \softmax(-\tfrac{1}{2\sigma^2}(r_1^2 + r_2^2))$. 
Finally, after yet another $\mulop$, we obtain
\[
H^{(8)} = 
\begin{bmatrix}
x_1 & \tilde y_1 & x_2 & \tilde y_2 & x_3 \\
x_1 & x_1 & x_2 & x_2 & x_3  \\ 
0 & y_1 & 0 & y_2 & 0  \\
0 &r_1 & 0 & r_2  & 
r_3  \\
0 &r_1^{2} & 0 & r_2^2  & 
-\tfrac{1}{2\sigma^2}(r_1^2 + r_2^2)  \\ 
0 & 0 & 0 & 0 & p \\ 
0 & 0 & 0 & 0 & p \circ r_3 
\end{bmatrix}.
\]
Above, $\circ$ denotes elementwise multiplication. 
Finally, after an $\affineop$ operation, we obtain 
\[
H^{(9)} = 
\begin{bmatrix}
x_1 & \tilde y_1 & x_2 & \tilde y_2 & x_3 \\
x_1 & x_1 & x_2 & x_2 & x_3  \\ 
0 & y_1 & 0 & y_2 & 0  \\
0 &r_1 & 0 & r_2  & 
r_3  \\
0 &r_1^{2} & 0 & r_2^2  & 
-\tfrac{1}{2\sigma^2}(r_1^2 + r_2^2)  \\ 
0 & 0 & 0 & 0 & p \\ 
0 & 0 & 0 & 0 & p \circ r_3 \\ 
0 & 0 & 0 & 0 & f^\star_\pi(P)
\end{bmatrix}.
\]
Note that the bottom-right entry of $H^{(9)}$ contains
the output $f^\star_\pi(P)$. In other words, the desired 
result if $H^{(9)}_{2d + 4m + 2, 2k + 1}$. Note that 
$H \in \R^{p \times q}$ where $p = 2 \dimension + 4 \numcomponents + 
2$, and $q = 2 \promptlen + 1$.
\subsection{Proof of Proposition~\ref{prop:implementable-ops}}
\label{sec:proof-prop-implementable-ops}
To begin with, we recall a few definitions, introduced in 
recent work~\citep{MaEtAl2023}. 

\bde [$\RAW$ operator]
The Read-Arithmetic-Write ($\RAW$) operators are maps on 
matrices, $\R^{p \times q} \to \R^{p \times q}$, 
\[
\RAW_{\bullet}(H; \cI, \cJ, \cK, \Theta_\cI, \Theta_\cJ, \Theta_\cK, \pi) = H', 
\quad \mbox{where}~\quad \bullet \in \{\otimes, \oplus\}. 
\]
Here $\pi$ is a causal set-valued map, with $\pi(i) \subset [i-1]$. 
The operators $\otimes, \oplus$ denote elementwise multiplication and addition, respectively. 
The entries of $H'$ are given by 
\begin{subequations}
\begin{align}
H'_{\cK, i} &\defn \Theta_\cK \bigg( 
\Theta_\cJ H_{\cJ, i} \bullet 
\Big(
\frac{\Theta_\cI}{\max\{|\pi(i)|,1\}} 
\sum_{i' \in \pi(i)} H_{\cI, i'}
\Big)
\bigg), \quad \mbox{and} \\ 
H'_{\cK^c, i} &= H_{\cK^c, i}, 
\end{align}
\end{subequations}
for each $i \in [q]$. Note that above $\cK^c = [p]~\setminus~\cK$, and for some 
positive integer $r$, $\Theta_\cI \in \R^{r \times |\cI|}, 
\Theta_\cJ \in \R^{r \times |\cJ|}$, and $\Theta_\cK \in \R^{|\cK| \times r}$. 
\ede 

In~\citet{MaEtAl2023}, they show that the $\RAW$ operator can be implemented in one autoregressive transformer layer. They also argue that (with a slight change in parameterization) 
that the $\mulop$ and $\affineop$ operators are transformer-implementable. Therefore, we simply need to argue that the operators $\softmaxop, \copydown, \copyover$, and $\scaledagg$
are all implementable by a transformer. 

To begin with, note that, by inspection, we have, 
with $\delta = \ell - k$,
\begin{subequations}
\label{subeq:RAW-reduction}
    \begin{align}
&\copyover(H; k, k', \ell, \cI) = 
\RAW_\oplus(H; [k, \ell], 
\emptyset, [k', k' + \delta], I_{\delta + 1},
0, I_{\delta + 1}, \pi_{\cI}) \\ 
&\copydown(H; k, k', \ell, \cI) = 
\RAW_\oplus(H;[k, \ell], 
\emptyset, [k', k' + \delta], I_{\delta+ 1},
0, I_{\delta + 1}, \pi'_\cI) \\ 
&\scaledagg(H; \alpha, k, \ell, k', i, \cI) 
= 
\RAW_\oplus(H; [k, \ell], 
\emptyset, [k', k' + \delta], I_{\delta + 1},
0, \alpha I_{\delta + 1}, \pi''_{\cI, i})
    \end{align}
\end{subequations}
Above, note that the intervals $[a, b]$ 
are just the integers between $a$ and $b$ (inclusive) and 
that we have defined
\[
\pi_\cI(i) = \begin{cases} 
\{i-1\} & i \geq 2 \\
\emptyset & \text{otherwise}
\end{cases}, 
\quad 
\pi'_\cI(i) = \begin{cases} 
\{i\} & i \in \cI \\
\emptyset & \text{otherwise}
\end{cases}, 
\quad 
\mbox{and} \quad 
\pi''_{\cI, i}(j) = \begin{cases} 
\cI & j = i  \\
\emptyset & \text{otherwise}
\end{cases}.
\] 
Therefore, the displays~\eqref{subeq:RAW-reduction} 
establish the following result. 
\ble 
The operators $\copyover, \copydown,$ and $\scaledagg$
are all implementable via the $\RAW$ operator in a 
single autoregressive transformer layer. 
\ele 

Finally, in Section~\ref{sec:proof-of-softmax} we 
demonstrate the following result. 
\ble\label{lem:softmax}
The softmax operation is implementable 
by an autoregressive transformer. 
\ele 

This completes the proof of 
Proposition~\ref{prop:implementable-ops}.

\subsubsection{Proof of Lemma~\ref{lem:softmax}}
\label{sec:proof-of-softmax}
In order to implement the softmax operation, we need to introduce a few 
other operations: 
\begin{itemize}
\item $\divop(H; j, k, k', \ell, \cI)$: \quad For columns with index $i \in \cI$, outputs $H'$ where $H'_{k' + t, i} = H_{k + t, i} / H_{j, i}$
for all $t \in \{0, \dots, \ell - k\}$. The remaining 
entries of $H$ are copies as is into $H'$.
\item $\movop(H; k, k', \ell, \cI)$: \quad For columns with index 
$i \in \cI$, outputs $H'$ where $H'_{k' + t, i} = H_{k + t, i}$
for all $t \in \{0, \dots, \ell - k\}$. 
 The remaining 
entries of $H$ are copies as is into $H'$.
\item $\sigmoid(H; k, k'):$ \quad In the final column 
$q$, outputs $H'$ with $H'_{k', q} = \frac{1}{1 + \e^{-H_{k , q}}}$.
The remaining entries of $H$ are copies as is into $H'$.
\end{itemize}

The operations $\divop, \movop$ are special cases of the 
same operations as introduced in the paper~\citet{MaEtAl2023}. 
Thus, we only need to demonstrate that $\sigmoid$ is transformer-implementable.
Assuming this for the moment, note that 
the softmax operation $\softmax$ is then implementable by the 
following sequence of operations. Let $H$ denote the input 
to the softmax layer, and let $s = H_{k:\ell, q}$. 
Using the affine operation ($\affineop$) together with the softmax operation 
($\softmaxop$) we can compute the values $1/(1 + \e^{s_i})$. 
Using the affine operation ($\affineop$) together with the $\divop$ operation, 
we can invert these values to compute $\e^{s_i}$. 
Finally, we can compute the sum of these values $S = \sum_i \e^{s_i}$ with 
an affine operation ($\affineop$) and we can divide by this sum 
using another $\divop$ operation. The result values are $\e^{s_i}/ 
\sum_{j} \e^{s_j}$, which is the softmax of the vector $s$. 
A move operation ($\movop$) then can move these values into the 
correct locations, $H_{k':(k' + \ell - k), q}$, as required. 

Thus, to complete the proof, we need to show how to implement the sigmoid 
operation. For this, we can begin by using the affine operation to insert
a value of $1$ in the final column, and another 
affine operation to insert a $2 \times 2$ identity matrix in the first 
$2$ columns of $H$. Then by selecting 
$\KeyMat$ and $\QueryMat$ to select the identity matrix and 
to select $(H_{k, i}, 1)$, respectively, we can 
ensure that $\KeyMat H_{:i}$ is a $2 \times i - 1$ matrix 
with alternating columns $(1, 0), (0, 1)$. We also have 
$\QueryMat h_i = (H_{k, i}, 1)$. The corresponding 
softmax values in the self-attention layer are $s_i = 
(\e^{H_{k + t, i}}/(1+H_{k, i}), 1/(\e^{H_{k, i}}  + 1))$. 
By selecting $\ValueMat$ to select the identity matrix in 
$H_{:i}$, and $\CombineMat$ to select the first value of $s_i$ and 
place it in position $k' + t$, we can ensure that 
$a =1/(\e^{-H_{k, i}}  + 1) e_{k'}$, where $e_j$ denotes the $j$th standard 
basis vector. 
This value is precisely the sigmoid, as needed. 
To place this value in the correct location, we simply set 
the feedforward network matrices $\FFInMat, \FFOutMat = 0$. 
Then, to preserve the output, we need to delete the value $1$, and
identity matrices placed into $H$ at the beginning; this can 
clearly be done by two affine operations. 

\section{Additional details on training methodology}
\label{sec:training-methodology}
Our training approach closely follows that of~\cite{garg2022can} and~\cite{MaEtAl2023}. 
After some hyperparameter optimization, we settled on the choice of hidden dimension of $256$, $8$ attention heads, and $12$ layers. We trained our transformers using Adam, with a constant step size of $0.1$. We used curriculum training, as in~\cite{garg2022can}, with the exception of Figure~\ref{fig:sample-size}, where 
the sample size was fixed. Our curriculum phases were $2000$ steps each, 
with a batch size of $64$. The final stage of training had $250000$ steps with $64$ batches. 
All of our figures presented mean squared errors computed over batch sizes of $256$. 
The dimension of the original covariates was $d =20$ throughout this paper. 

\subsection{Details on fixed sample-size training}
\label{sec:fixed-sample-size-training}

In this setting, we used hyperparameter tuning over the dropout parameter, 
$\rho \in \{0, 0.05, 0.1\}$, and found the following choices to be best, for Figure~\ref{fig:sample-size}: 
\begin{itemize}
    \item for $n = 15000$, we took $\rho = 0.0$.
    \item for $n = 30000$, we took $\rho = 0.1$.
    \item for $n = 45000$, we took $\rho = 0.0$.
    \item for $n = 60000$, we took $\rho = 0.1$.
\end{itemize}
We also used curriculum training in this setup, but obtained the samples 
by subsampling the fixed dataset. This was done by first randomly 
sampling a batch from the full dataset, and then randomly dropping 
and shuffling the prefix of each prompt so as to obtain a prompt of the 
shorter, desired length. Otherwise, the entire procedure was the same 
as the other figures, as described above. 

\subsection{Batch expectation maximization (EM) algorithm}
\label{sec:EM}
Batch expectation-maximization is a variant of the standard expectation-maximization method (see, for instance, Section 14.5.1 in \citet{Bishop06}). For completeness, we 
describe the algorithm formally here. Note that 
$\phi$ denotes the standard univariate Gaussian pdf below. For notation, 
we also denote the prompts as 
\[
P^{(i)} \defn 
\big(x^{(i)}_1, y^{(i)}_1, \dots, 
x^{(i)}_\promptlen, y^{(i)}_\promptlen, 
x^{(i)}_{\promptlen +1}\big), 
\quad \mbox{for}~i \in [n].
\]
The algorithm is then stated below as Algorithm~\ref{alg:EM} 
\begin{algorithm}[H]
\caption{Batch expectation-maximization for a discrete mixture of 
linear regressions with Gaussian noise}
\label{alg:EM}
\begin{algorithmic}
\Require Length $\promptlen$ prompts $\{P^{(i)}\}_{i=1}^n$  
noise variance $\sigma > 0$, 
number of components $\numcomponents > 0$. 
\State \emph{Initialize} $\pi^{(0)} \in [0, 1]^\numcomponents$, drawn 
uniformly on the probability simplex. 
\State \emph{Initialize} $w_j^{(0)} \in \R^\dimension$, drawn uniformly on the 
sphere of radius $\sqrt{\dimension}$ for $j \in [\numcomponents]$.
\State \emph{Initialize} $\gamma_{ij}^{(0)} = 0$ for all $i \in [n], 
j \in [\numcomponents]$.
\While{have not converged}
\State update prompt-component assignment probabilities,
\[
\gamma_{ij}^{(t + 1)} =
\frac{\pi_{j}^{(t)} 
\prod_{l=1}^\promptlen 
\phi\Big(\frac{y^{(i)}_l - 
(x^{(i)}_l)^\T w_j^{(t)}}{\sigma}\Big)
}{
\sum_{j' = 1}^{\numcomponents} 
\pi_{j'}^{(t)}
\prod_{l=1}^\promptlen 
\phi\Big(\frac{y^{(i)}_l - 
(x^{(i)}_l)^\T w_{j'}^{(t)}}{\sigma}\Big)
}, 
\quad \mbox{for all}~i \in [n], j \in 
[\numcomponents].
\]
\State update the marginal component probabilities by the formula 
\[
\pi^{(t + 1)}_j = \frac{1}{n} 
\sum_{i=1}^n \gamma_{ij}^{(t+1)}, 
\quad \mbox{for all}~j \in [\numcomponents]
\]
\State update the parameter estimates by solving,
\[
w_j^{(t+1)} = 
\argmin_{w \in \R^\dimension} 
\Big\{
\sum_{i=1}^n \sum_{l=1}^\promptlen 
\gamma_{ij}^{(t + 1)} 
\big(y^{(i)}_l - w^\T x^{(i)}_l\big)^2
\Big\}, \quad \mbox{for all}~j \in [\numcomponents].
\]
\State update the iteration counter, $t \gets t + 1$.
\EndWhile \\
\Return final set of component centers, $\{w_j^{(t)}\}_{j =1}^\numcomponents$
\end{algorithmic}
\end{algorithm}

In our implementation we stop (or declare the algorithm converged) if 
$t > t_{\rm max}$, or if 
\[
\max_{j} \min_{j'} \|w_j^{(t)} - w_{j'}^{(t-1)}\|_2 \leq \varepsilon. 
\]
In our experiments we took $t_{\rm max} = 20000$ and $\varepsilon = 0.001$. 

\section{Comparison to distribution shift 
with the posterior mean estimator}
\label{sec:comparison-to-pma-on-shfit}

In this section, we replicate the figures presented in Section~\ref{sec:distribution-shift}, except we evaluate the distribution shift settings on the posterior mean procedure, $f^\star_\pi$ as defined in display~\eqref{eqn:posterior-mean-alg}.
\begin{figure}[H]
\includegraphics[width=0.49\linewidth]{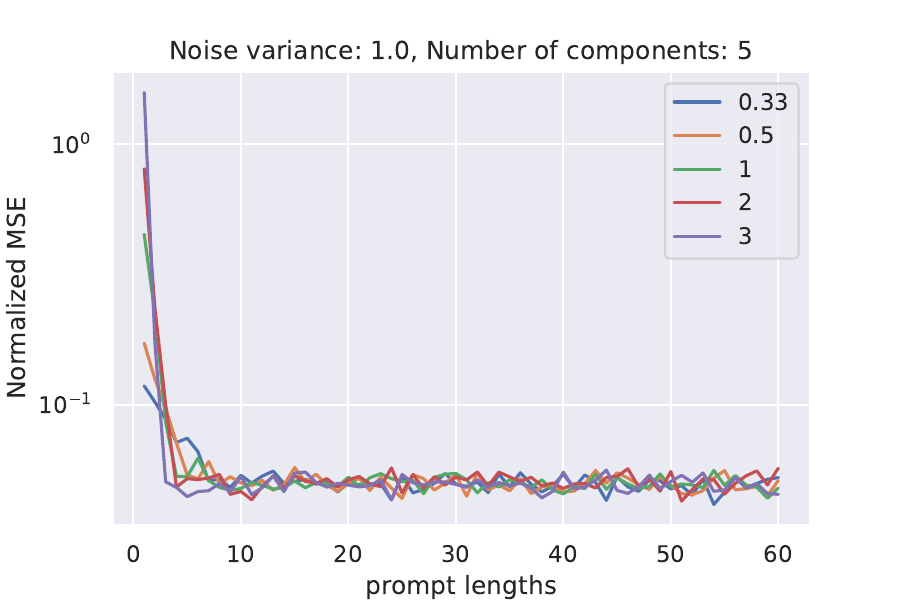}
\includegraphics[width=0.49\linewidth]{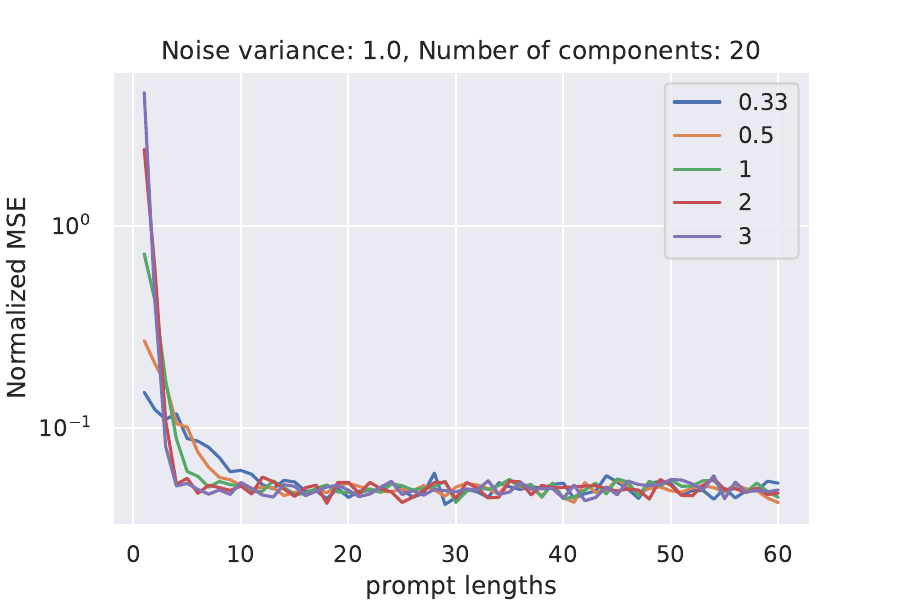}\\
\includegraphics[width=0.49\linewidth]{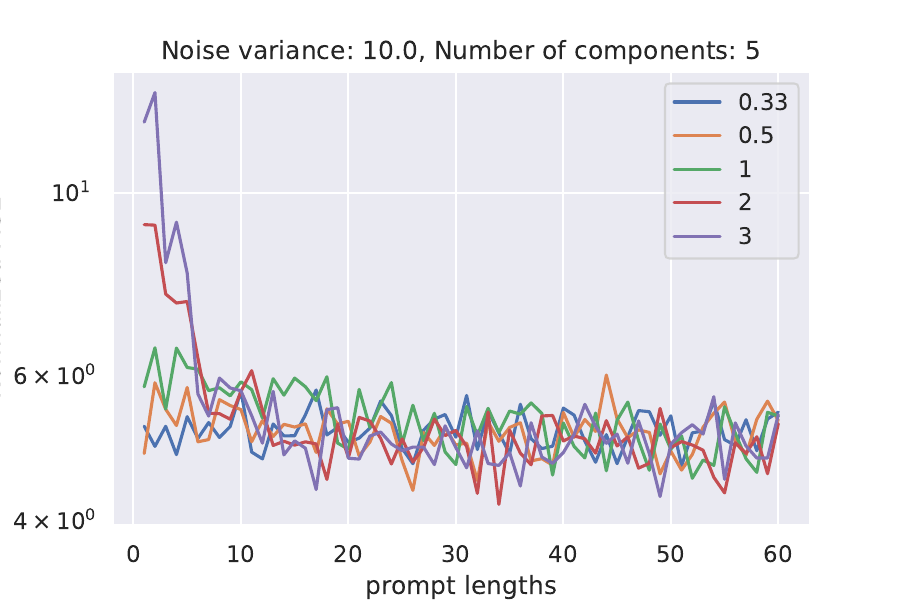}
\includegraphics[width=0.49\linewidth]{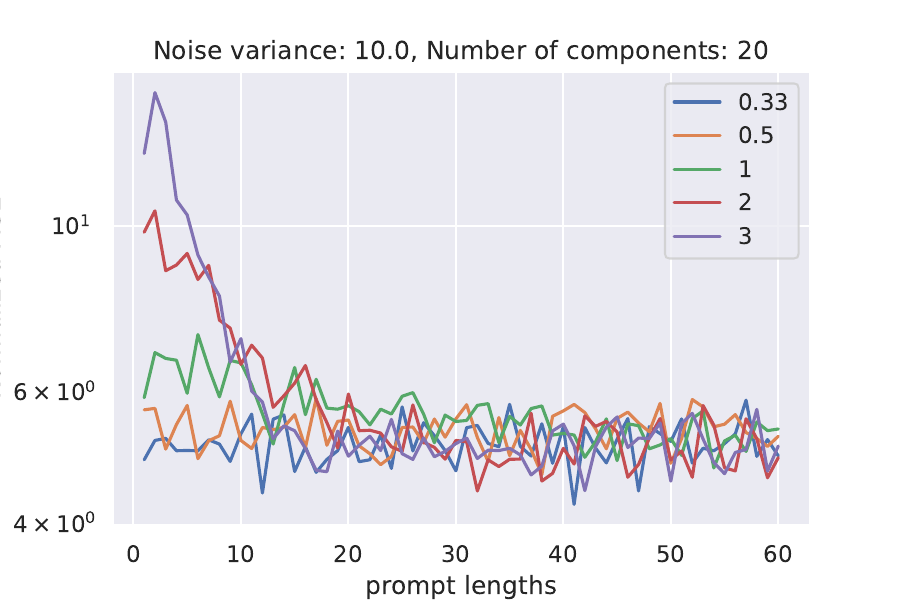}
\caption{Posterior mean algorithm on covariate scaling distribution shift setting.}
\end{figure} 
\begin{figure}[H]
\includegraphics[width=0.49\linewidth]{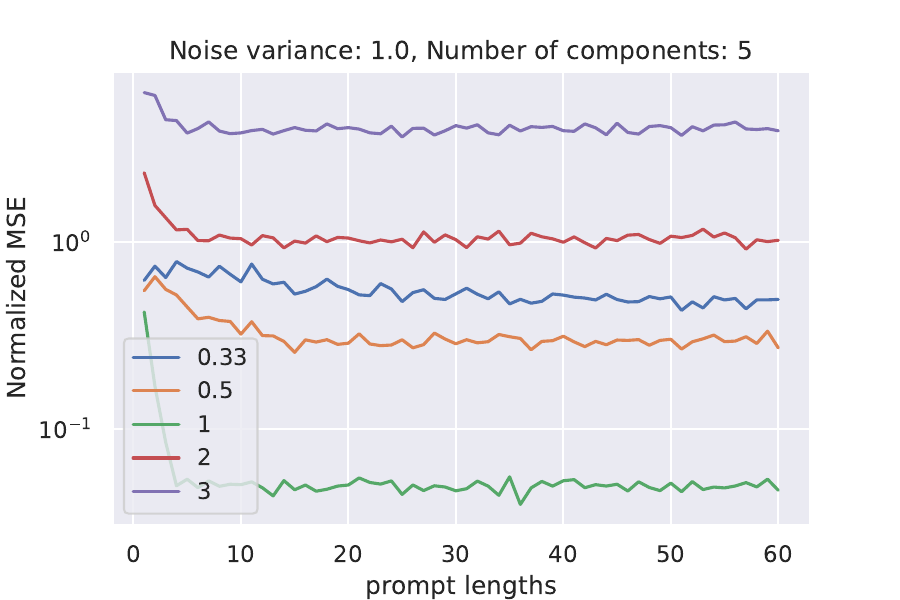}
\includegraphics[width=0.49\linewidth]{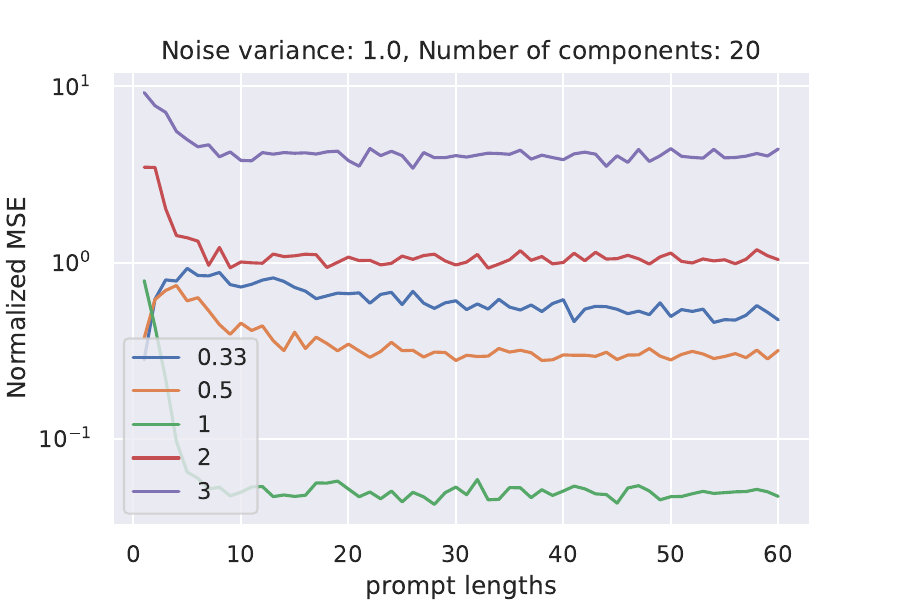}\\
\includegraphics[width=0.49\linewidth]{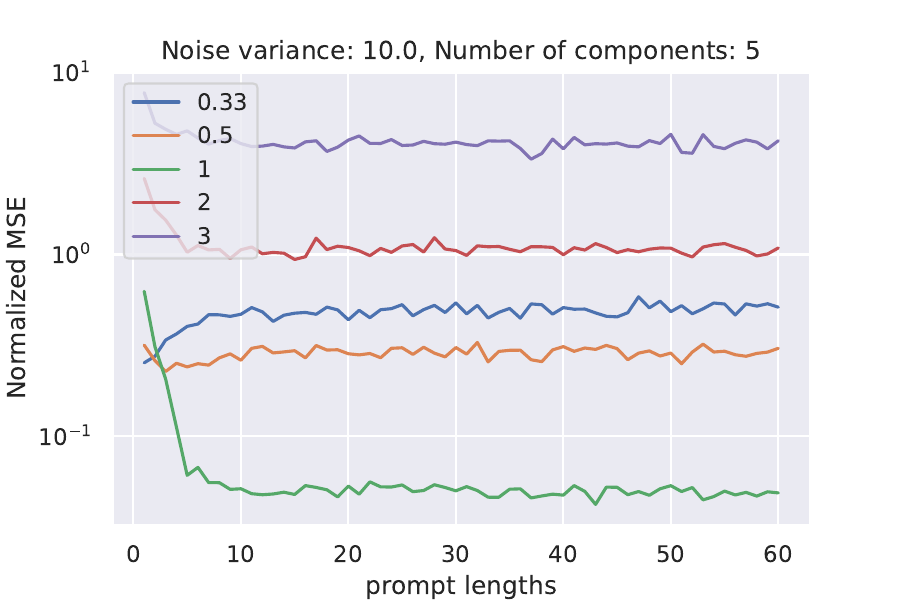}
\includegraphics[width=0.49\linewidth]{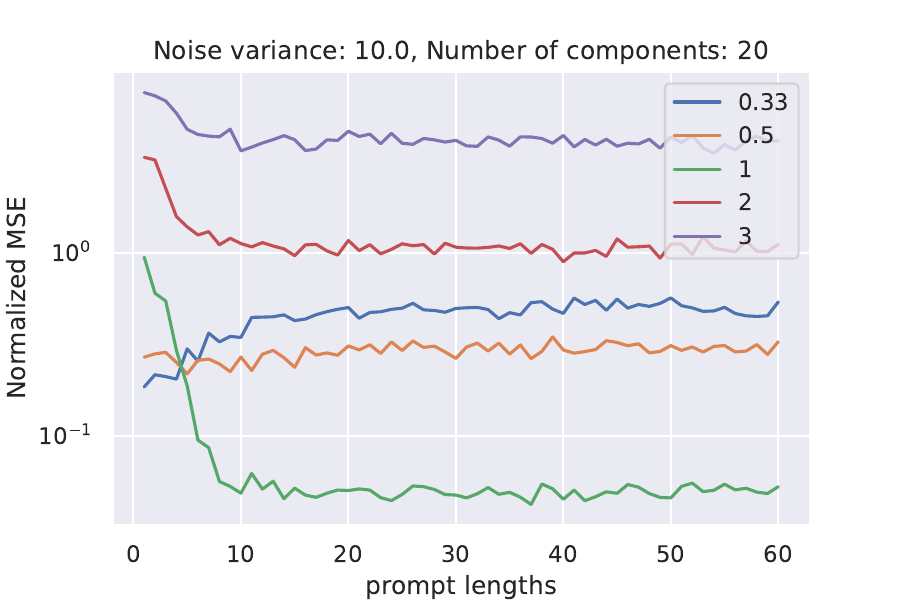}
\caption{Posterior mean algorithm on weight scaling distribution shift setting.}
\end{figure} 
\begin{figure}[H]
\includegraphics[width=0.49\linewidth]{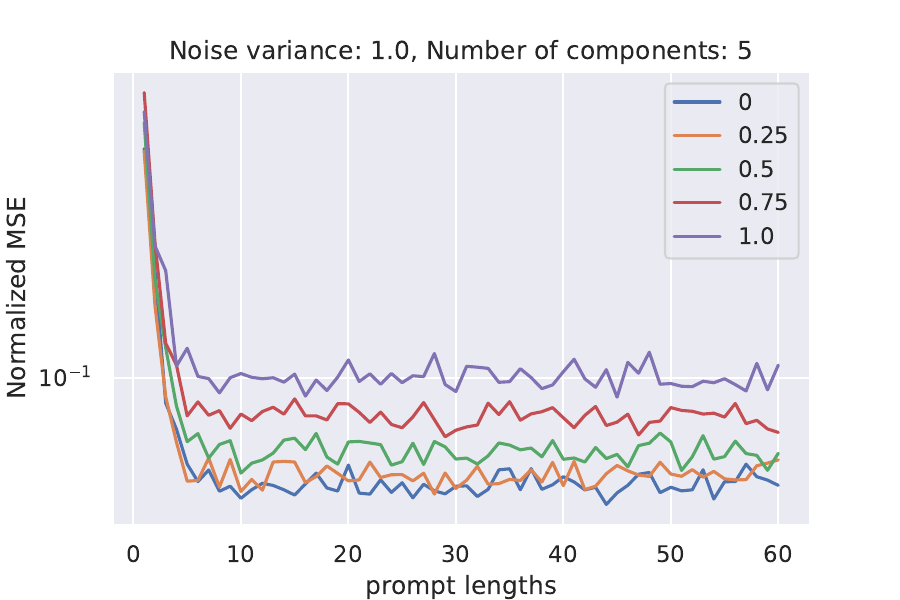}
\includegraphics[width=0.49\linewidth]{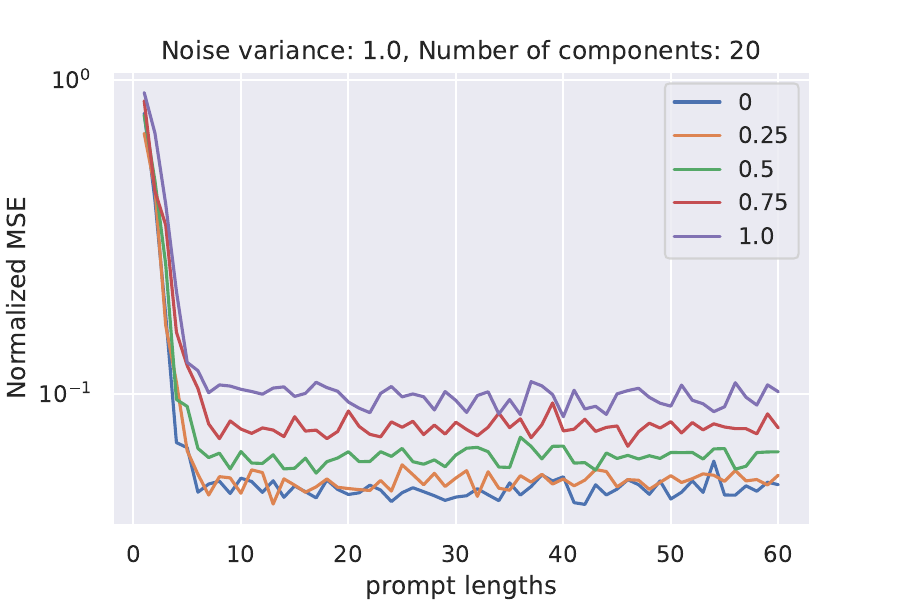}\\
\includegraphics[width=0.49\linewidth]{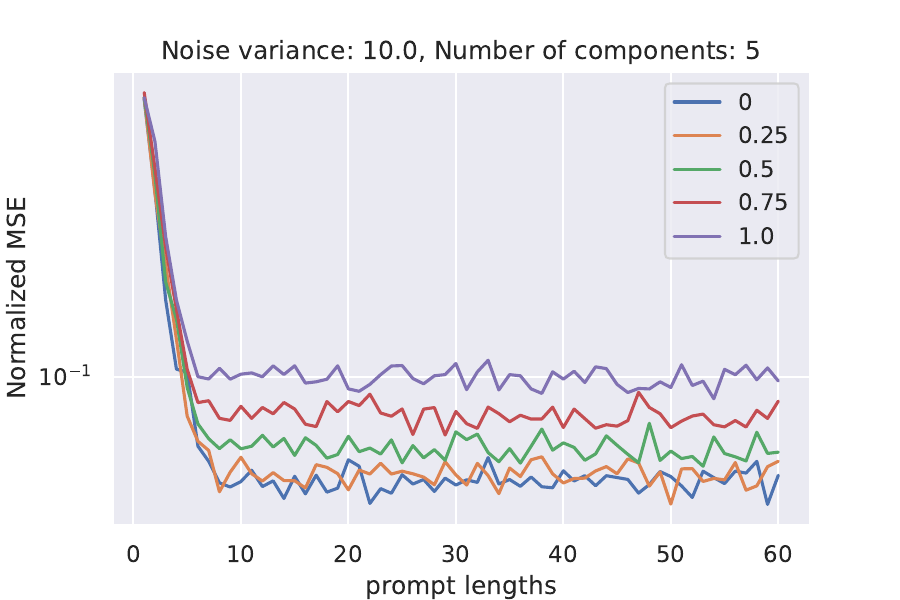}
\includegraphics[width=0.49\linewidth]{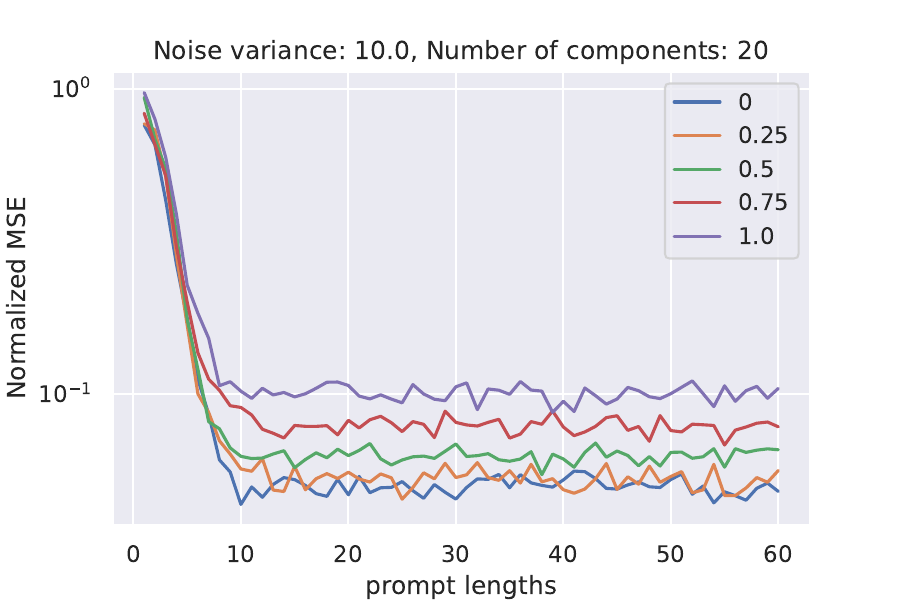}
\caption{Posterior mean algorithm on weight additive shift setting.}
\end{figure} 
\end{document}